\documentclass{article}

\PassOptionsToPackage{authoryear,round}{natbib}
\PassOptionsToPackage{table}{xcolor}
\usepackage[preprint]{neurips_2025}

\usepackage[utf8]{inputenc}
\usepackage[T1]{fontenc}
\usepackage{hyperref}
\usepackage{url}
\usepackage{booktabs}
\usepackage{amsmath,amsfonts,amssymb}
\usepackage{nicefrac}
\usepackage{microtype}
\usepackage{xcolor}
\usepackage{graphicx}
\usepackage{colortbl}
\usepackage{makecell}
\usepackage{tabularx}
\usepackage{array}
\usepackage{placeins}
\usepackage{dblfloatfix}
\usepackage{caption}
\usepackage{ragged2e}
\usepackage[most]{tcolorbox}
\usepackage{longtable}
\usepackage{listings}
\usepackage{upquote}
\tcbuselibrary{listings,breakable,skins}

\definecolor{wgaccent}{HTML}{175CD3}
\definecolor{wgframe}{HTML}{D0D5DD}
\definecolor{wgback}{HTML}{F9FAFB}
\definecolor{wgtitle}{HTML}{123B67}
\definecolor{wghead}{HTML}{F2F4F7}

\lstdefinestyle{wgprompt}{
  basicstyle=\ttfamily\scriptsize,
  breaklines=true,
  breakatwhitespace=false,
  columns=fullflexible,
  keepspaces=true,
  showstringspaces=false,
  upquote=true,
  tabsize=2,
  aboveskip=0pt,
  belowskip=0pt,
  frame=none,
  keywordstyle=\color{wgaccent}\bfseries,
  alsoletter={_},
  keywords={INSTRUCTION,OCCUPATION_PLAYBOOK,LINEAGE_MEMORY,SYNTHESIS_REQUEST,ANCHOR_MANIFEST,ANCHOR_FILES,REFERENCE_FILE_MANIFEST,REFERENCE_FILES,REFERENCE_FILES_PARSED,TASK_PROMPT,RUBRIC,UNSATISFIED_ITEMS,REFERENCE_DELIVERABLE_PARSED,DEFECT_SET,ROUND_INDEX,WORKSPACE_DIR,ADDITIONAL_TOOLS,MAX_SEARCH_ATTEMPTS,REQUIRED_FILE_FORMATS,DELIVERABLE_AGENT_CAPABILITIES,SOLVER_TOOLS,RUBRIC_SIZE_TARGET,ARTIFACT_CHECK_REPORT,MEDIA_HANDLING_NOTES,LINEAGE_MEMORY_FILE}
}

\newtcblisting{wgpromptbox}[1][]{
  breakable,
  enhanced,
  listing only,
  listing options={style=wgprompt},
  colback=wgback,
  colframe=wgframe,
  coltitle=wgtitle,
  colbacktitle=wghead,
  boxrule=0.6pt,
  arc=1.5mm,
  left=2mm,
  right=2mm,
  top=1.5mm,
  bottom=1.5mm,
  fonttitle=\bfseries\small,
  title={#1}
}

\definecolor{bbblue}{HTML}{175CD3}
\definecolor{bbink}{HTML}{123B67}
\definecolor{flashgray}{HTML}{667085}
\definecolor{caseband}{HTML}{F2F4F7}

\hypersetup{
  colorlinks=true,
  citecolor=neuripsaccentblue,
  linkcolor=neuripsaccentblue,
  urlcolor=neuripsaccentblue
}

\title{WorkGenesis: Building the Worlds That Teach Agents to Work}
\author{%
Xinyu Zhu\textsuperscript{1,2,*}\quad
Fenyi Liu\textsuperscript{1,2,*}\quad
Yuzhu Cai\textsuperscript{2,3}\quad
Shuo Tang\textsuperscript{1}\quad
Rui Ye\textsuperscript{1}\quad
Linfeng Zhang\textsuperscript{1,2}\\
Siheng Chen\textsuperscript{1,2,\ensuremath{\dagger}}\\[0.3em]
\textsuperscript{1}Shanghai Jiao Tong University\quad
\textsuperscript{2}Endless Frontier\quad
\textsuperscript{3}Beihang University\\[0.2em]
{\small
\textsuperscript{*}Equal contribution.\quad
\textsuperscript{\ensuremath{\dagger}}Corresponding author.}
}

\begin{document}

\maketitle

\begin{abstract}
The ability of Large Language Model (LLM) agents to complete daily and professional work is receiving increasing attention. Training such agents requires realistic work scenarios. Expert-authored occupational work is costly and slow to produce, while unconstrained synthesis often yields tasks with weak factual grounding or internally inconsistent requirements. To bridge this gap, we introduce \textbf{WorkGenesis}, a framework that constructs executable occupational work from real-world artifacts through two core technical innovations: (1) \textit{Evidence-Based Work Construction}, which grounds each unit of work in real-world evidence by retrieving public files guided by O*NET occupational knowledge and synthesizing the surrounding context, companion materials, work request, and itemwise rubric around them; and (2) \textit{Execution-Guided Consistency Verification}, which renders a reference deliverable inside the constructed work, attributes every unsatisfied rubric item to the agent, the task, or the rubric, and uses task and rubric defects as feedback to iteratively repair the work until it passes the audit. Experimental results demonstrate that \textbf{Fx-Work-35B}, trained with simple supervised fine-tuning (SFT) on only 20K units of work synthesized by WorkGenesis, achieves the highest scores among all comparable-scale baselines on the five reported metrics across GDPvalAA-v2, APEX-Agents-AA, and JobBench (31.00 versus 24.79 average score), and even surpasses frontier models such as the 1.6T DeepSeek-V4-Pro-Preview. These results show that WorkGenesis provides scalable training data for working agents.
\begin{flushleft}
\begin{tabular}{@{}ll@{}}
\includegraphics[width=1em]{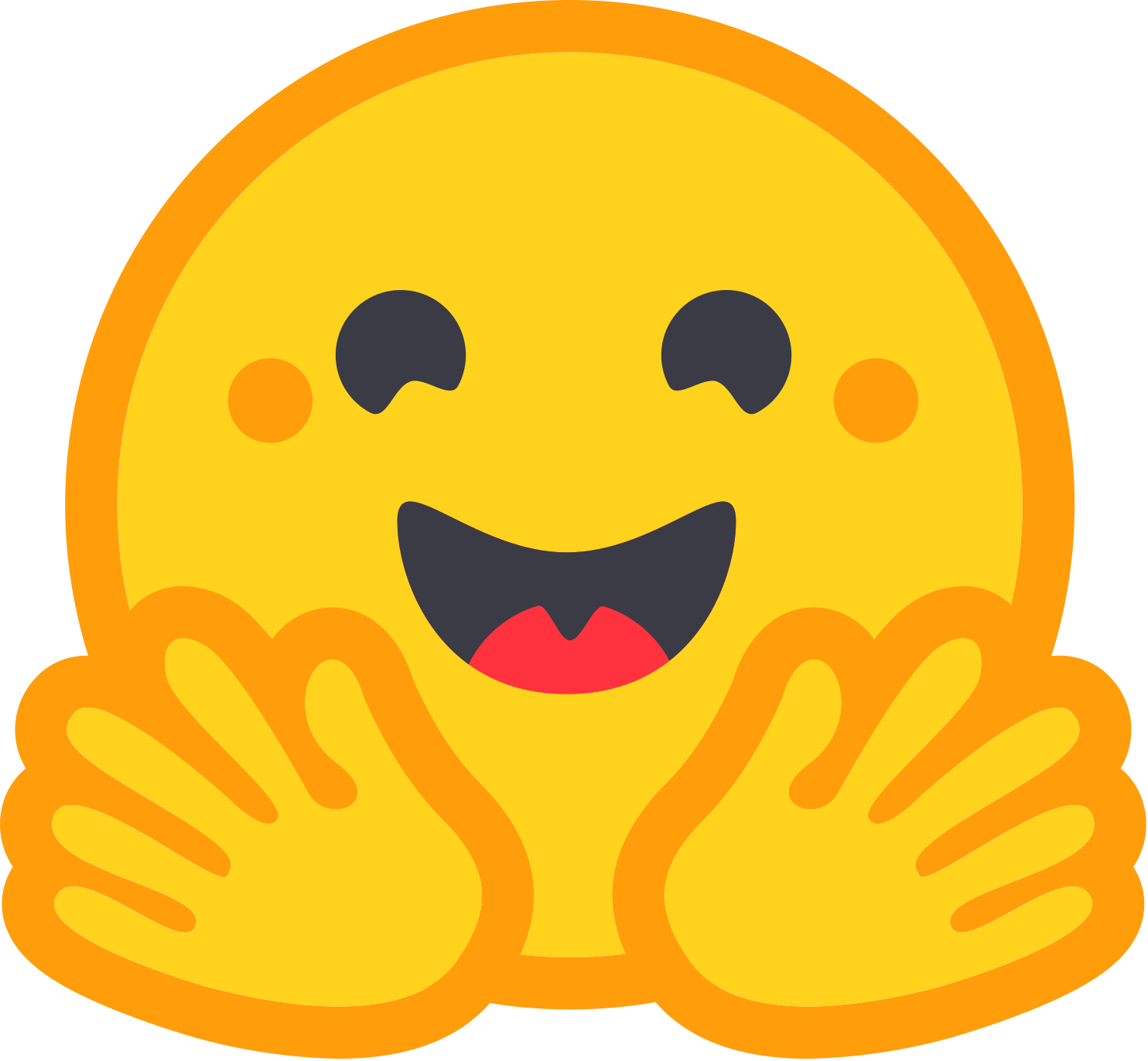}
\quad \textbf{Model} &
\href{https://huggingface.co/endless-frontier/Fx-Work}
     {https://huggingface.co/endless-frontier/Fx-Work}
\end{tabular}
\end{flushleft}
\end{abstract}

\begin{figure}[!h]
    \centering      
    \includegraphics[width=1\linewidth]{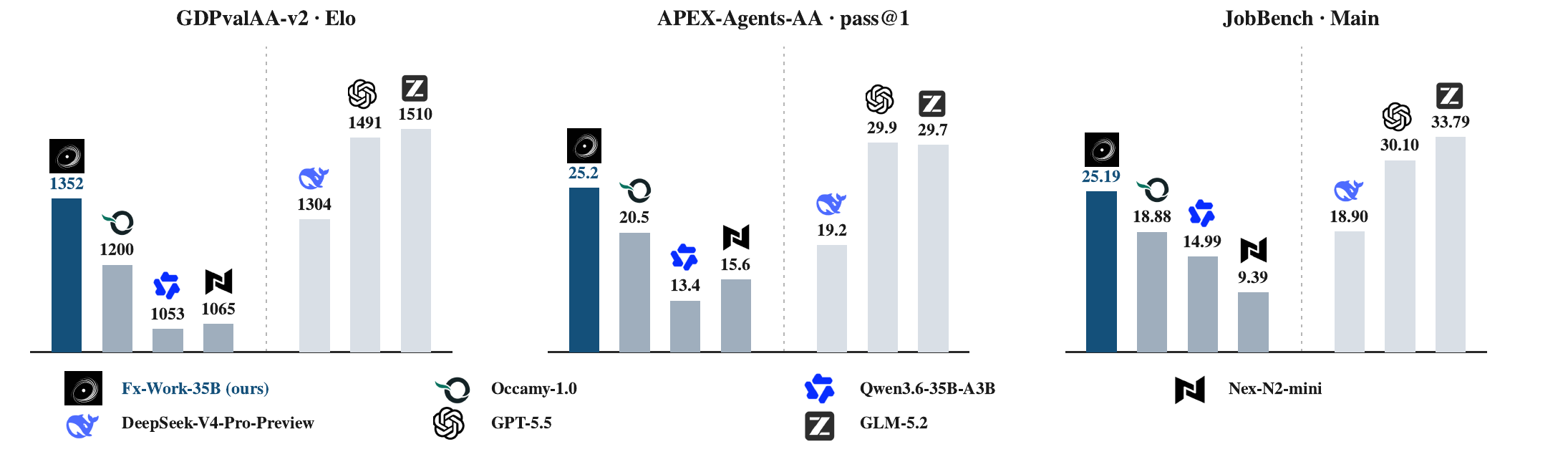}
    \caption{Performance comparison of different models on three authoritative occupational work benchmarks. Fx-Work-35B, trained  on work scenarios synthesized by WorkGenesis, outperforms all comparable-scale models as well as DeepSeek-V4-Pro-Preview with 1.6T parameters.} 
    \label{fig:main_result} 
\end{figure}

\section{Introduction}
\label{sec:introduction}

Large Language Model (LLM) agents are moving from answering questions toward completing work in environments. Advances in tool use and long-horizon reasoning enable language models to search, compute, manipulate files, and produce deliverables that can be assessed against professional standards \citep{openai2025gdpval,jobbench2026}. Research on environment synthesis and simulation\citep{zuo2026qwenagentworld} is also bringing the environments themselves into the design space of agent learning. Workplace intelligence gives this trajectory a concrete objective: agents must organize knowledge and action amid heterogeneous materials, operational constraints, and intermediate feedback to complete meaningful professional activities.

However, environments that can support occupational learning remain difficult to construct at scale. A credible workplace setting requires interconnected materials, a coherent business context, and defensible delivery requirements alongside its occupational role and task description. High-fidelity occupational evaluations rely on experienced professionals to author tasks and on multiple rounds of review \citep{openai2025gdpval}. Extending this instance-by-instance process into a training-data supply demands sustained, costly professional effort. Traditional data synthesis also cannot solve this problem. Without constraints from real materials, a setting may offer little beyond occupational language; without execution-based scrutiny, fluent descriptions can conceal contradictions among evidence, objectives, and evaluation criteria. The scarce resource is professional work that supports action and withstands examination.

To address this gap, we introduce \textbf{WorkGenesis}, a framework that turns underused records of real work into \textbf{executable occupational work}. Real work leaves records, including documents, intermediate analyses, templates, and other files, that preserve facts about the setting, constraints, and conventions of a job. These records are abundant, but they are rarely organized into tasks for agent learning. WorkGenesis treats them as evidence from which to assemble a complete work scenario: it supplies missing context and materials, states an actionable objective, and defines what a satisfactory deliverable must contain. Grounded in occupational knowledge from O*NET \citep{onetdatabase}, this process connects a fragmentary record to a professional activity that an agent can perform and that can be evaluated.

 WorkGenesis has two complementary components. \textbf{Evidence-Based Work Construction} uses real records as evidence and organizes them with occupational knowledge into a complete scenario with context, materials, objectives, and acceptance criteria. \textbf{Execution-Guided Consistency Verification} then runs that scenario and uses the resulting deliverable to assess whether the work is feasible from the supplied information, professionally reasonable, internally consistent, and appropriately difficult. If execution exposes a problem in the scenario or its evaluation criteria, the construction and verification process repairs it before acceptance. Together, these components turn underused records into work that an agent can perform and that can be evaluated.



To assess the training utility of this work, we construct \textbf{20K} units of occupational work and collect their tool-interaction trajectories. Supervised fine-tuning (SFT) of Qwen3.6-35B-A3B yields \textbf{Fx-Work-35B}. We evaluate the resulting model from three perspectives: economically valuable work through GDPval, long-horizon cross-application work through APEX-Agents, and work that professionals prioritize for delegation through JobBench \citep{openai2025gdpval,vidgen2026apex,jobbench2026}. Under our GDPvalAA-v2 protocol, Elo increases from \textbf{1053 to 1352}; APEX-Agents pass@1 rises from \textbf{13.4\% to 25.2\%}; and JobBench Main and Easy scores rise from \textbf{14.99 / 56.02} to \textbf{25.19 / 70.06}, respectively. WorkGenesis achieves the highest scores on all five reported metrics among the comparable-scale baselines in our evaluation. These results support the central proposition of this paper: occupational work built from general knowledge and grounded in real materials can provide useful training data for economically valuable and human-desired professional activities.

Our contributions can be summarized as follows:
\begin{itemize}
    \item \textbf{Automated Construction of High-Quality Occupational Work.} We introduce WorkGenesis, which uses Evidence-Based Work Construction and Execution-Guided Consistency Verification to transform real-world files and occupational knowledge into executable work.
    \item \textbf{Training and Empirical Validation of Fx-Work-35B.} We train Fx-Work-35B on 20K occupational tasks synthesized by WorkGenesis. Fx-Work-35B significantly outperforms comparable-scale models on three authoritative occupational-task benchmarks, even surpassing DeepSeek-V4-Pro-Preview, demonstrating the utility of the occupational scenarios synthesized by WorkGenesis.
    \item \textbf{Detailed Analysis and Open-Source Release.} We provide a multi-dimensional analysis of the distribution, diversity, and other properties of the WorkGenesis-synthesized occupational task data, as well as the performance and cost-effectiveness of Fx-Work-35B. We open-source Fx-Work-35B, hoping to advance the development of working agents.
\end{itemize}

\section{Related Work}
\label{sec:related_work}

\textbf{Benchmarks for professional and computer-use agents.}
Recent benchmarks have moved agent evaluation from static question answering to work performed in realistic environments. WorkArena~\citep{drouin2024workarena} and OSWorld~\citep{xie2024osworld} evaluate browser and computer-use agents on multi-step tasks, while TheAgentCompany~\citep{xu2025theagentcompany}, CRMArena~\citep{huang2025crmarena}, and $\tau$-bench~\citep{yao2024taubench} study consequential work and tool-agent-user interaction in particular domains. GDPval~\citep{patwardhan2025gdpval} measures economically valuable deliverables, APEX-Agents~\citep{vidgen2026apex} stresses long-horizon professional workflows, and JobBench~\citep{jobbench2026} focuses on work that people want to delegate. These benchmarks are important targets for progress, but their primary purpose is evaluation; they do not by themselves provide a scalable way to construct diverse, file-backed training scenarios with auditable requirements.

\textbf{Synthetic data and environment construction.}
Instruction synthesis methods such as Self-Instruct~\citep{wang2023selfinstruct}, WizardLM~\citep{xu2024wizardlm}, PersonaHub~\citep{chan2024personahub}, and Magpie~\citep{xu2025magpie} show that language models can expand training data from compact seeds. For agents, DreamGym~\citep{chen2026dreamgym} and related environment-synthesis efforts such as Echoverse~\citep{pandya2026echoverse} and Qwen-AgentWorld~\citep{zuo2026qwenagentworld} generate experiences or interactive worlds at scale. WorkGenesis addresses a complementary gap: it begins with public artifacts as factual anchors, uses occupational knowledge to organize them into a workplace scenario, and produces a request, companion files, and rubric that jointly define an executable deliverable. This evidence-first construction keeps the scenario connected to real work while allowing the surrounding materials to be generated automatically.

\textbf{Verification and iterative improvement.}
Self-Refine~\citep{madaan2023selfrefine} and Reflexion~\citep{shinn2023reflexion} use feedback to improve an agent's answer or subsequent behavior, and execution-based checking has been used to expose defects in evaluation data, as illustrated by SciCode-Verified~\citep{scicodeverified}. WorkGenesis applies execution as a consistency test for the work scenario itself. A reference agent attempts the task, every unsatisfied rubric item is attributed to the agent, the task, or the rubric, and defects traced to the latter two are repaired before the scenario is accepted. Feedback thus checks whether the prompt, reference files, and evaluation criteria describe the same professional activity rather than polishing a single response.

\section{Method}
\label{sec:method}

\subsection{Overview and Problem Formulation}
\label{sec:overview}

To make agents capable on daily professional work, we need to construct professional work at scale that is realistic, executable, and self-checkable. We formulate a unit of professional work as

\begin{equation}
\label{eq:work}
\mathcal{W}=(q,\mathcal{F},\mathcal{R}),\qquad \mathcal{F}=\mathcal{F}^{\mathrm{real}}\cup\mathcal{F}^{\mathrm{syn}}.
\end{equation}

Here $q$ is a work request that states the occupational role, business context, objective, and delivery format; $\mathcal{F}$ is the set of reference files the work uses, comprising real public files $\mathcal{F}^{\mathrm{real}}$ and synthesized companion attachments $\mathcal{F}^{\mathrm{syn}}$; and $\mathcal{R}$ is the rubric, a set of itemwise acceptance criteria for the deliverable. Together they turn a professional situation into a file-backed unit of work: an agent reads the reference files, completes the task, and produces a deliverable that can be evaluated item by item against $\mathcal{R}$.

Constructing such work requires solving two problems. (1) \textbf{Task plausibility.} Occupational tasks written directly by large language models often appear plausible, yet they are disconnected from real working scenarios and highly homogeneous in task style. Real working scenarios are grounded in actual documents and the professional context around them, which is exactly what free generation lacks. (2) \textbf{Task consistency.} Professional work usually involves multiple complex requirements and a large number of reference files. The task requirements, the reference files, and the final evaluation rubric must therefore agree with one another, so that the work does not reward an agent for meeting a requirement that contradicts its own materials or for satisfying a criterion that the request never implies.

To solve these two problems, we introduce two mechanisms. (1) \textbf{Evidence-Based Work Construction} (\S\ref{sec:construction}). We leverage occupational information from O*NET to retrieve occupation-related documents from the public web as real-world anchors. On these anchors we construct the occupational task scenario and its evaluation rubric, and then have an agent attempt the task inside that scenario to produce a reference deliverable. (2) \textbf{Execution-Guided Consistency Verification} (\S\ref{sec:repair}). We score the deliverable against the rubric, identify every item that does not receive full credit, and attribute each failing item to the rubric or to the task itself. We then use these items as a feedback signal to revise the task scenario until no such issue remains.

\begin{figure}[!h]
\centering
\includegraphics[width=1\linewidth]{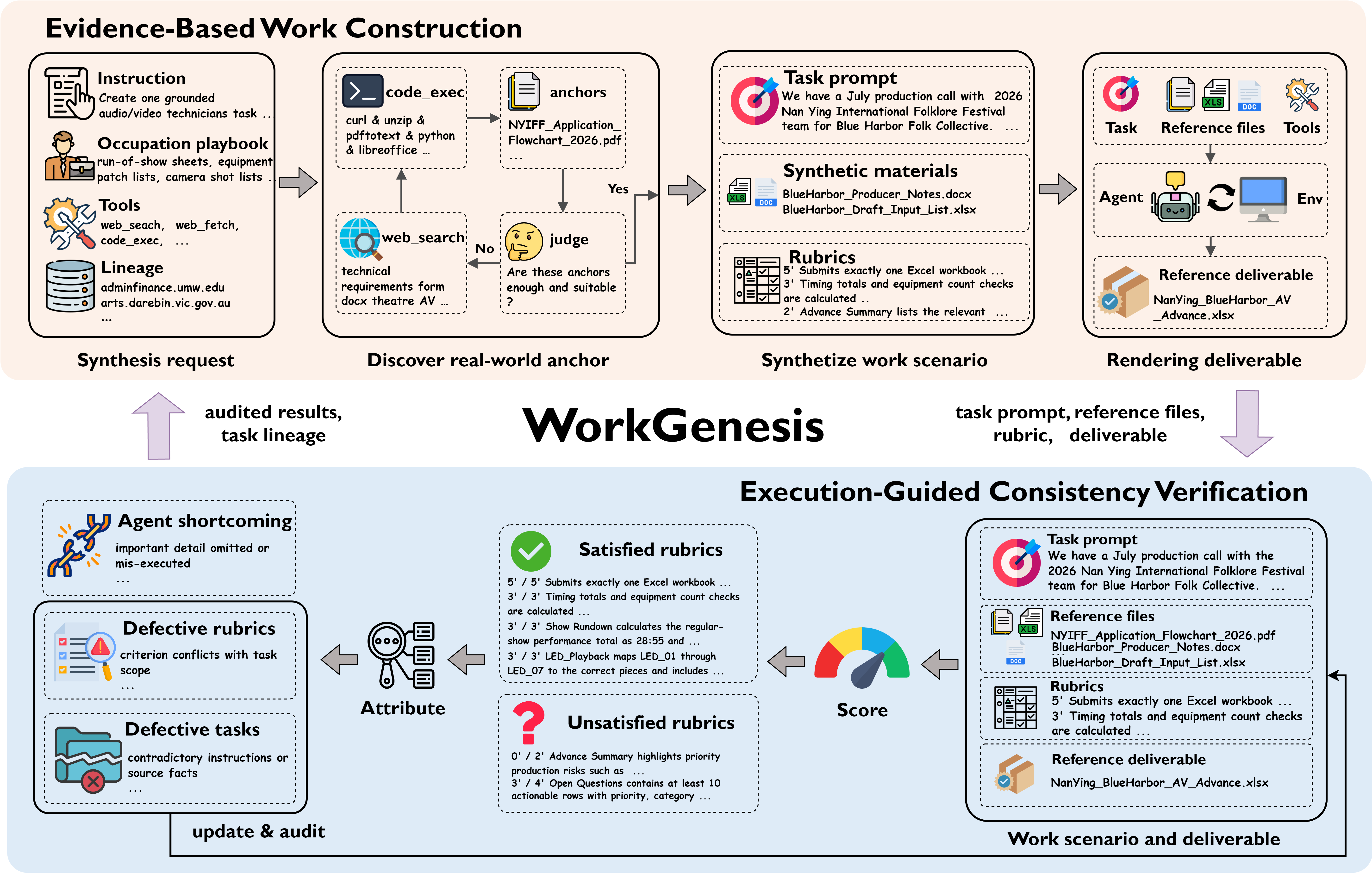}
\caption{Overview of the WorkGenesis framework. (a) \textbf{Evidence-Based Work Construction}: WorkGenesis repeatedly searches for suitable real-world anchor files and synthesizes additional reference files from them to construct the complete work scenario. It then assigns an independent agent to operate in this scenario and produce a reference deliverable. (b) \textbf{Execution-Guided Consistency Verification}: WorkGenesis scores the reference deliverable using a rubric and attributes unsatisfied rubric items, identifying defects in both the synthesized occupational task itself and the rubric. These defects are used as feedback signals to further revise the entire work scenario. This verification process repeats until no rubric item is attributed to the occupational scenario itself.}
\label{fig:pipeline}
\end{figure}

\subsection{Evidence-Based Work Construction}
\label{sec:construction}

We introduce a pipeline that constructs occupational work from real-world evidence. The pipeline takes a synthesis request as input and returns one complete unit of work together with the reference deliverable it asks for in four stages: reading the synthesis request, discovering real-world anchors, synthesizing the work scenario, and rendering the deliverable.

\textbf{Synthesis request.}
Every construction starts from a request with four parts, $c=(\iota,\omega,\mathcal{T},\mathcal{M})$. The instruction $\iota$ states the construction order and the behavioral requirements (e.g. \emph{Create one grounded audio/video technicians task}). The occupation playbook $\omega$ carries occupational information and duties from O*NET, which steers both the document search and the shape of the task. The tool set $\mathcal{T}$ comprises what the agent may use during construction such as web search and code execution. The lineage memory $\mathcal{M}$ keeps what earlier tasks in the same lineage have already consumed (e.g. source URLs and SHA-256 hashes of the real-world anchor files), so that a new task neither reuses an anchor file nor repeats a similar scenario.

\textbf{Discovering real-world anchors.}
To keep a constructed task plausible and aligned with real working scenarios, every unit of work depends on files that exist in the real world. Starting from $c$, WorkGenesis searches the public web for candidates that fit the occupation and the format named in the request, downloads and parses them with code execution, and judges whether a candidate is suitable as the foundation of a professional task. Retrieval and judgment alternate until a usable set of anchors is found:
\begin{equation}
\label{eq:anchors}
\varnothing\neq\mathcal{F}^{\mathrm{real}}\subseteq\mathcal{P}_c,\qquad \mathcal{F}^{\mathrm{real}}\cap\mathcal{M}=\varnothing,
\end{equation}
where $\mathcal{P}_c$ is the candidate pool retrieved under $c$. The second condition rules out files that an earlier task already used, which keeps anchors, and therefore scenarios, diverse.

\textbf{Synthesizing the work scenario.}
An anchor alone rarely supports a complex professional task. WorkGenesis therefore synthesizes companion attachments around the anchors to complete the information the task needs, keeps them together with the anchors as reference files, and then generates the work request and its rubric:
\begin{equation}
\label{eq:scenario}
\big(q,\mathcal{F}^{\mathrm{syn}},\mathcal{R}\big)=\mathrm{Construct}\big(c,\mathcal{F}^{\mathrm{real}}\big),\qquad \mathcal{W}_0=\big(q,\mathcal{F}^{\mathrm{real}}\cup\mathcal{F}^{\mathrm{syn}},\mathcal{R}\big).
\end{equation}
Here $\mathrm{Construct}(\cdot)$ denotes the construction pipeline, which synthesizes the companion attachments from the anchors and then the request and the rubric from the resulting materials together with the occupational information in $c$. The request $q$ is written as an ordinary workplace request and states the role, business context, objective, audience, constraints, and delivery format. The rubric $\mathcal{R}$ states how the deliverable will be judged. It contains dozens of criteria with different weights, covering content, computation, structure, and compliance, and each criterion is decidable from the request or the reference files. The request, the reference files, and the rubric together form the work scenario $\mathcal{W}_0$ that is to be examined.

\textbf{Rendering the deliverable.}
WorkGenesis then hands the scenario to an independent agent and asks it to do the work. The agent sees only the task prompt, the reference files, and the given tools, and it interacts with the environment until it submits a deliverable:
\begin{equation}
\label{eq:deliverable}
y^{\mathrm{ref}}\sim p_{\mathrm{agent}}\big(\cdot\mid\mathcal{W}_0;\mathcal{T}\big),
\end{equation}
where $p_{\mathrm{agent}}$ is the execution agent's policy over tool interactions and submissions, and $y^{\mathrm{ref}}$ is the reference deliverable it returns. Since this agent sees exactly what a future worker would see, $y^{\mathrm{ref}}$ is a fair probe of the scenario. It shows whether the reference files actually support the request, whether the request points to a well-defined deliverable, and whether the rubric can be applied item by item.

\subsection{Execution-Guided Consistency Verification}
\label{sec:repair}

A scenario can be internally inconsistent even when each part of it looks reasonable on its own. The request may refer to contents that the reference files do not contain, a rubric item may conflict with the task scope, or a criterion may demand more than the materials can support. Defects of this kind are invisible when the task is written and become visible only when the work is actually attempted. WorkGenesis therefore treats the reference deliverable as evidence about the scenario, verifies and repairs the scenario until that evidence stops revealing defects.

\textbf{Scoring the deliverable.}
The audit proceeds in passes indexed by $k=0,1,\dots$, and $\mathcal{W}_k=(q_k,\mathcal{F}^{\mathrm{real}}\cup\mathcal{F}^{\mathrm{syn}}_k,\mathcal{R}_k)$ denotes the scenario at the start of pass $k$, with request $q_k$, synthesized attachments $\mathcal{F}^{\mathrm{syn}}_k$, and rubric $\mathcal{R}_k$; pass $0$ examines the initially constructed scenario $\mathcal{W}_0$. At pass $k$, an LLM evaluator scores the reference deliverable against the rubric item by item. Each criterion is judged on its own, with the parsed deliverable, the reference files, and the task prompt attached to the request so that the judgment can be traced back to the materials. The items that do not receive full credit are collected into the substandard set $\Delta_k$:
\begin{equation}
\label{eq:substandard}
\Delta_k=\big\{(r_{k,j},s_{k,j}):r_{k,j}\in\mathcal{R}_k,\ s_{k,j}<s_{k,j}^{\max}\big\},
\end{equation}
where $r_{k,j}$ is the $j$-th criterion of the rubric $\mathcal{R}_k$ at pass $k$, $s_{k,j}$ is the score the reference deliverable receives on it, and $s_{k,j}^{\max}$ is that criterion's full-credit value.

\textbf{Attributing the failure.}
A criterion that is not fully met has more than one possible cause. The reference deliverable may have failed a reasonable requirement, the criterion may itself be defective, or the scenario may contradict itself. WorkGenesis attributes every item in $\Delta_k$ to one or more of three causes,
\begin{equation}
\label{eq:attribution}
\mathcal{C}=\{\mathsf{Agent},\mathsf{Rubric},\mathsf{Task}\},\qquad \lambda_k:\Delta_k\longrightarrow 2^{\mathcal{C}}\setminus\{\varnothing\},
\end{equation}
where $\mathcal{C}$ is the cause set and $\lambda_k$ is the attribution map at pass $k$, which assigns every substandard item at least one label. $\mathsf{Agent}$ marks an item that a reasonable requirement is not met because the agent lacked the ability to meet it. $\mathsf{Rubric}$ marks a criterion that conflicts with the reference files, misreads the request, or is defined too strictly. $\mathsf{Task}$ marks a scenario that is inconsistent with itself, for instance a prompt that refers to contents the reference files do not provide. A criterion may receive several labels, which keeps the overlapping responsibilities that occur in real work.

\textbf{Update and re-audit.}
Items attributed to $\mathsf{Rubric}$ or $\mathsf{Task}$ form the defect set $\Delta_k^{\mathrm{def}}=\{d\in\Delta_k:\lambda_k(d)\cap\{\mathsf{Rubric},\mathsf{Task}\}\neq\varnothing\}$ and serve as the feedback signal for revision. WorkGenesis then updates the task prompt, the reference files, and the rubric with the repair operator $\Phi$:
\begin{equation}
\label{eq:repair}
\mathcal{W}_{k+1}=\Phi\big(\mathcal{W}_k;\Delta_k^{\mathrm{def}}\big)=\big(q_{k+1},\mathcal{F}^{\mathrm{real}}\cup\mathcal{F}^{\mathrm{syn}}_{k+1},\mathcal{R}_{k+1}\big).
\end{equation}
Here $q_{k+1}$, $\mathcal{F}^{\mathrm{syn}}_{k+1}$, and $\mathcal{R}_{k+1}$ are the revised request, companion attachments, and rubric. The real anchors $\mathcal{F}^{\mathrm{real}}$ stay fixed, so a repair never rewrites the evidence on which the scenario is built; only the constructed parts change. Items attributed to $\mathsf{Agent}$ are recorded but left alone, since a capability gap in the reference agent is not a defect of the scenario. The resulting scenario $\mathcal{W}_{k+1}$ is rendered and scored again at the next pass, which places every revision back under the same test.

\textbf{Evolving and iterating.}
Scoring, attribution, verifying and repairing repeat until a pass yields no defective item, that is, until $\Delta_k^{\mathrm{def}}=\varnothing$, at which point the unit of work is accepted as successfully constructed. If the same criterion is still attributed to $\mathsf{Rubric}$ or $\mathsf{Task}$ after three rounds of verification, the scenario is judged to have a fundamental problem: WorkGenesis discards it and constructs a new task from the same initial synthesis request. Either way, the key facts of the run, such as source URLs and SHA-256 hashes, are written into the lineage memory $\mathcal{M}$, so that the next construction avoids both the same anchor files and similar scenarios.

\section{Experiments}
\label{sec:experiments}

\subsection{Experimental Setup}
\label{sec:exp_setup}
\textbf{Implementation.}
To validate the effectiveness of the occupational work constructed by WorkGenesis and its utility for improving agents' capability on professional tasks, we use GLM-5.2~\citep{zai2026glm52} as the backend LLM that drives the components of WorkGenesis. Starting from 65 occupations in O*NET~\citep{onetdatabase}, the pipeline constructs 20K units of occupational work, and we collect the complete trajectories of GLM-5.2 solving these tasks under the Stirrup agent harness, which yields 20K trajectories in total. During collection, we set the maximum context length to 256K tokens and trigger automatic compaction once 70\% of the context budget is consumed. Supervised fine-tuning on these trajectories over Qwen3.6-35B-A3B~\citep{qwen36_35b_a3b} yields Fx-Work-35B. We refer to the appendix for more implementation details.

\textbf{Benchmarks.}
We evaluate on the three benchmarks that measure the agent's capability of
professional work, including GDPval~\citep{patwardhan2025gdpval}, APEX-Agents~\citep{vidgen2026apex} and JobBench~\citep{jobbench2026}.
GDPval measures whether agents can complete economically valuable tasks in human society. APEX-Agents evaluates long-horizon, cross-application tasks created by investment banking analysts, management consultants, and corporate lawyers, which require agents to navigate realistic work environments with files and tools. JobBench measures the occupational work that people most want to delegate to AI. To ensure that the evaluation is accurate and comparable, we adopt the official public evaluation protocol of Artificial Analysis and use its official Stirrup harness~\citep{artificialanalysis2026gdpvalaa}. For GDPval, we report the normalized rubric score and the Elo rating; for APEX-Agents, the pass@1 score; and for JobBench, the scores on the main and easy splits. The average score for each model is computed by averaging the normalized Elo score on GDPvalAA-v2 and the normalized scores on APEX-Agents-AA and the JobBench Main split. Elo scores are normalized using the official formula, \((\mathrm{Elo} - 500) / 2000 \times 100\%\).

\textbf{Baselines.} To assess the efficacy of Fx-Work-35B, we compare it against a broad spectrum of state-of-the-art models categorized into two groups: (1) frontier large-size models, including Qwen3.8-Max~\citep{qwen38_max}, Kimi-K3~\citep{team2026kimi}, GPT-5.5~\citep{openai2026gpt55}, GLM-5.2~\citep{zai2026glm52} , DeepSeek-V4-Pro-Preview and DeepSeek-V4-Flash-Preview~\citep{deepseek2026v4}; (2) $\leq$ 35B models, which serve as direct, comparable-scale benchmarks. This group includes models released by various organizations and enterprises (e.g., Apodex-1.0-mini~\citep{apodex2026} and Agents-A1~\citep{bai2026agentsa1}), models specifically trained for professional tasks (e.g., Nex-N2-mini~\citep{nexagi2026nexn2} and Occamy-1.0~\citep{chen2026occamy10openparetofrontier35b}), and Qwen’s official baseline models, Qwen3.6-35B-A3B and Qwen3.6-27B~\citep{qwen36_35b_a3b}.

\begin{table*}[t]
\centering
\footnotesize
\caption{\textbf{Main results on GDPvalAA-v2, APEX-Agents-AA and JobBench.} Fx-Work-35B significantly outperforms all $\sim$35B-sized models on all three benchmarks. It even surpasses the 1.6T DeepSeek-V4-Pro-Preview in GDPval Elo score, as well as on APEX-Agents and JobBench. \textbf{Bold} marks the best value within the $\le$35B group. GDPval Elo scores are computed based on the August 4, 2026 snapshot and standard calculation methodology from Artificial Analysis. The average score for each model is computed by averaging the normalized Elo score on GDPvalAA-v2 and the normalized scores on APEX-Agents-AA and JobBench Main split.}
\label{tab:main}
\setlength{\tabcolsep}{5pt}
\renewcommand{\arraystretch}{1.2}
\begin{tabular}{llcccccc}
\toprule
& & \multicolumn{2}{c}{\textbf{GDPvalAA-v2}} & \multicolumn{1}{c}{\textbf{APEX-Agents-AA}} & \multicolumn{2}{c}{\textbf{JobBench}} & \\
\cmidrule(lr){3-4}\cmidrule(lr){5-5}\cmidrule(lr){6-7}
\textbf{Model} & \textbf{Params} & \textbf{Rubric} & \textbf{Elo}
& \textbf{pass@1} & \textbf{Main} & \textbf{Easy} & \textbf{AVG} \\
\midrule
\rowcolor{blue!13}\multicolumn{8}{c}{\emph{Frontier Models}}\\
Qwen3.8-Max            & 2.4T-A95B     & 91.53 & 1739 & 38.9 & 52.57 & 82.81 & 51.14  \\
Kimi-K3            & 2.8T-A104B    & 91.64 & 1687 & 35.5 & 43.81 & 80.72 & 46.22 \\
GPT-5.5            & --     & 90.50 & 1491 & 29.9 & 30.10 & 78.22 & 36.52 \\
GLM-5.2              & 753B-A40B     & 90.34 & 1510 & 29.7 & 33.79 & 75.49 & 37.80 \\
DeepSeek-V4-Pro-Preview     & 1.6T-A49B   & 87.17 & 1304 & 19.2 & 18.90 & 64.08 & 26.10 \\
DeepSeek-V4-Flash-Preview   & 284B-A13B   & 86.49 & 1189 & 15.2 & 18.62 & 63.13 & 22.75 \\
\midrule
\rowcolor{blue!13}\multicolumn{8}{c}{\emph{$\le$35B Models}}\\
Nex-N2-mini        & 35B-A3B    & 69.96 & 1065 & 15.6 & 9.39 & 52.93 & 17.74 \\
Agents-A1         & 35B-A3B    & 74.31 &  877 & 11.7 &  7.19 & 41.74 & 12.58 \\
Apodex-1.0-mini        & 35B-A3B   & 78.36 &  973 & 14.6 & 7.97 & 44.38 & 15.40 \\
Occamy-1.0        & 35B-A3B    & 79.50 &  1200 & 20.5 & 18.88 & 65.49 & 24.79 \\
Qwen3.6-27B         & 27B    & 85.10 & 1138 & 16.6 & 19.18 & 64.59 & 22.56 \\
Qwen3.6-35B-A3B 
                                           & 35B-A3B & 82.61 & 1053 & 13.4 & 14.99 & 56.02 & 18.68 \\
\rowcolor{caseband}
\textbf{Fx-Work-35B} \emph{(ours)}      & 35B-A3B
& \textbf{86.66} & \textbf{1352} & \textbf{25.2} & \textbf{25.19} & \textbf{70.06} & \textbf{31.00} \\
\bottomrule
\end{tabular}
\end{table*}

\begin{figure}[htbp]
\centering

\includegraphics[width=1\linewidth]{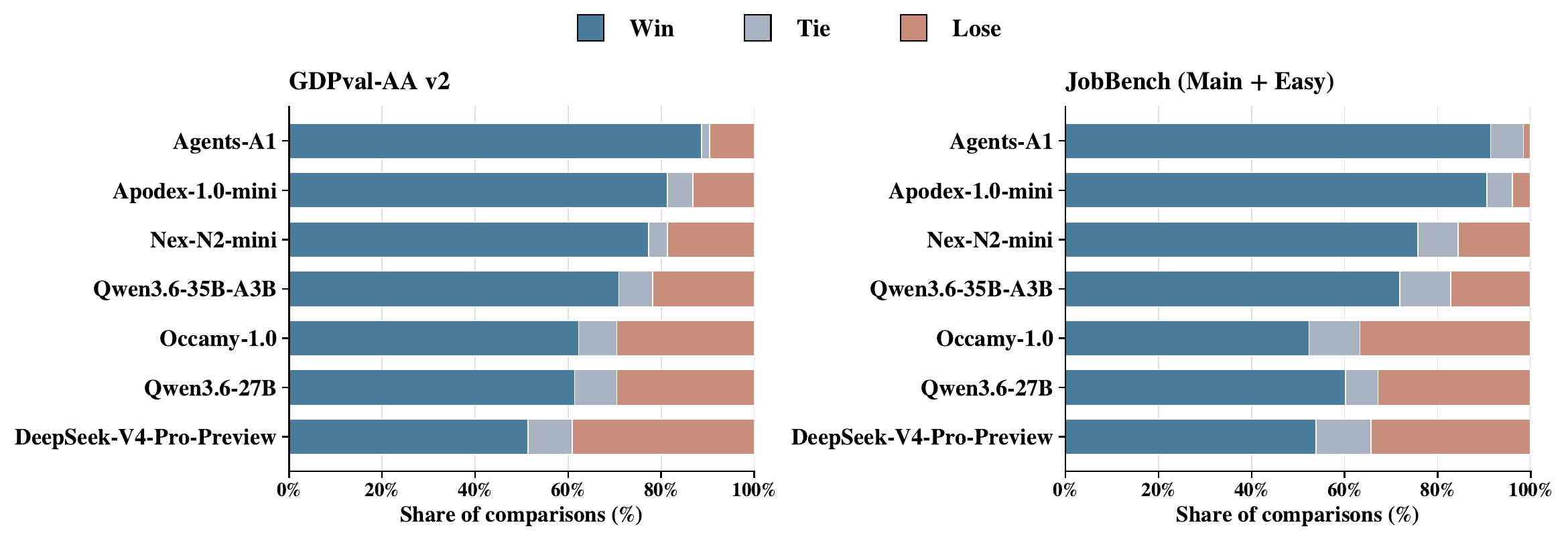}

\caption{Pairwise performance comparison of Fx-Work-35B and other baselines on GDPvalAA-v2 and JobBench. ``Win'' denotes the percentage of tasks on which Fx-Work-35B outperforms the compared model; ``Tie'' and ``Lose'' are defined analogously.}
\label{fig:pairwise}

\end{figure}

\textbf{WorkGenesis outperforms all comparable-scale models on three benchmarks.} As shown in Table~\ref{tab:main}, Fx-Work-35B surpasses all other comparable-scale models on three occupational work-task benchmarks including GDPvalAA-v2, APEX-Agents-AA, and JobBench, achieving state-of-the-art (SOTA) results. On GDPvalAA-v2, its rubric and Elo scores reach 86.66 and 1352, respectively, improving over the baseline by 4.05 and 299 points. On APEX-Agents-AA, it raises the task pass rate from 13.4\% at the baseline to 25.2\%. On the Main and Easy splits of JobBench, it achieves 25.19 and 70.06, respectively, improving over the baseline by 10.2 and 14.04. It also exceeds the strongest comparable-scale baseline on every metric, by 1.56 rubric points and 152 Elo points on GDPvalAA-v2, 4.7 pass@1 points on APEX-Agents, and 6.01 and 4.57 points on JobBench Main and Easy. With 35B parameters, the model surpasses the frontier model DeepSeek-V4-Pro-Preview at 1.6T on four of the five metrics and DeepSeek-V4-Flash-Preview at 284B on all five. Figure~\ref{fig:pairwise} reports the average win rate of Fx-Work-35B against other comparable-scale models on GDPvalAA-v2 and JobBench Main. Together, these results demonstrate that training on occupational tasks synthesized by WorkGenesis substantially improves Fx-Work-35B's ability to complete occupational work tasks.

\begin{figure}[t]
\centering
\begin{minipage}[t]{0.48\linewidth}
\centering
\includegraphics[width=\linewidth]{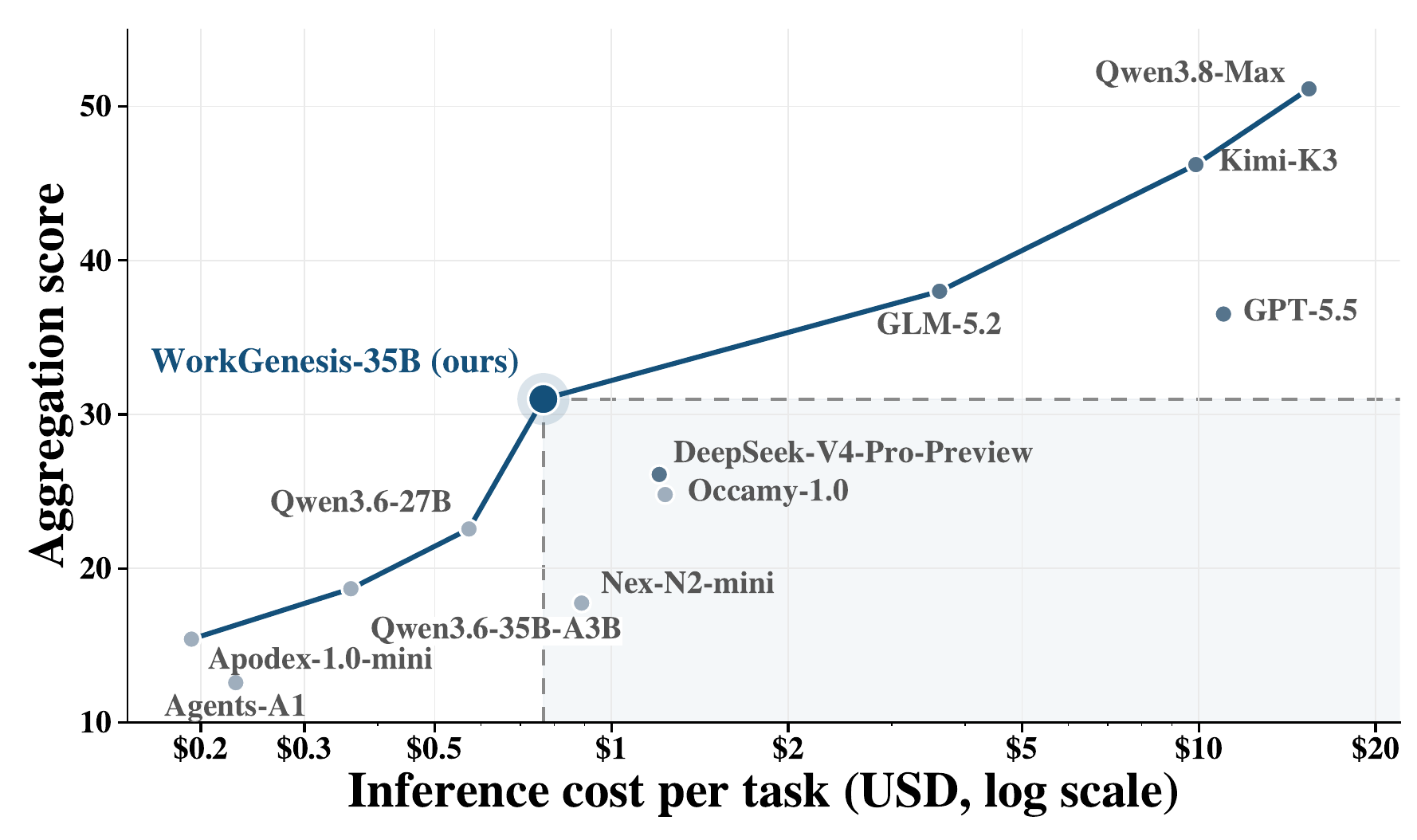}

\caption{\textbf{The cost and performance of different models.} The horizontal axis is the average cost of completing one task among three benchmarks and the vertical axis is the aggregated score among three benchmarks. Fx-Work-35B reaches the highest rating in the $\leq$35B group at a low cost per task.}
\label{fig:cost}
\end{minipage}\hfill
\begin{minipage}[t]{0.48\linewidth}
\centering
\includegraphics[width=\linewidth]{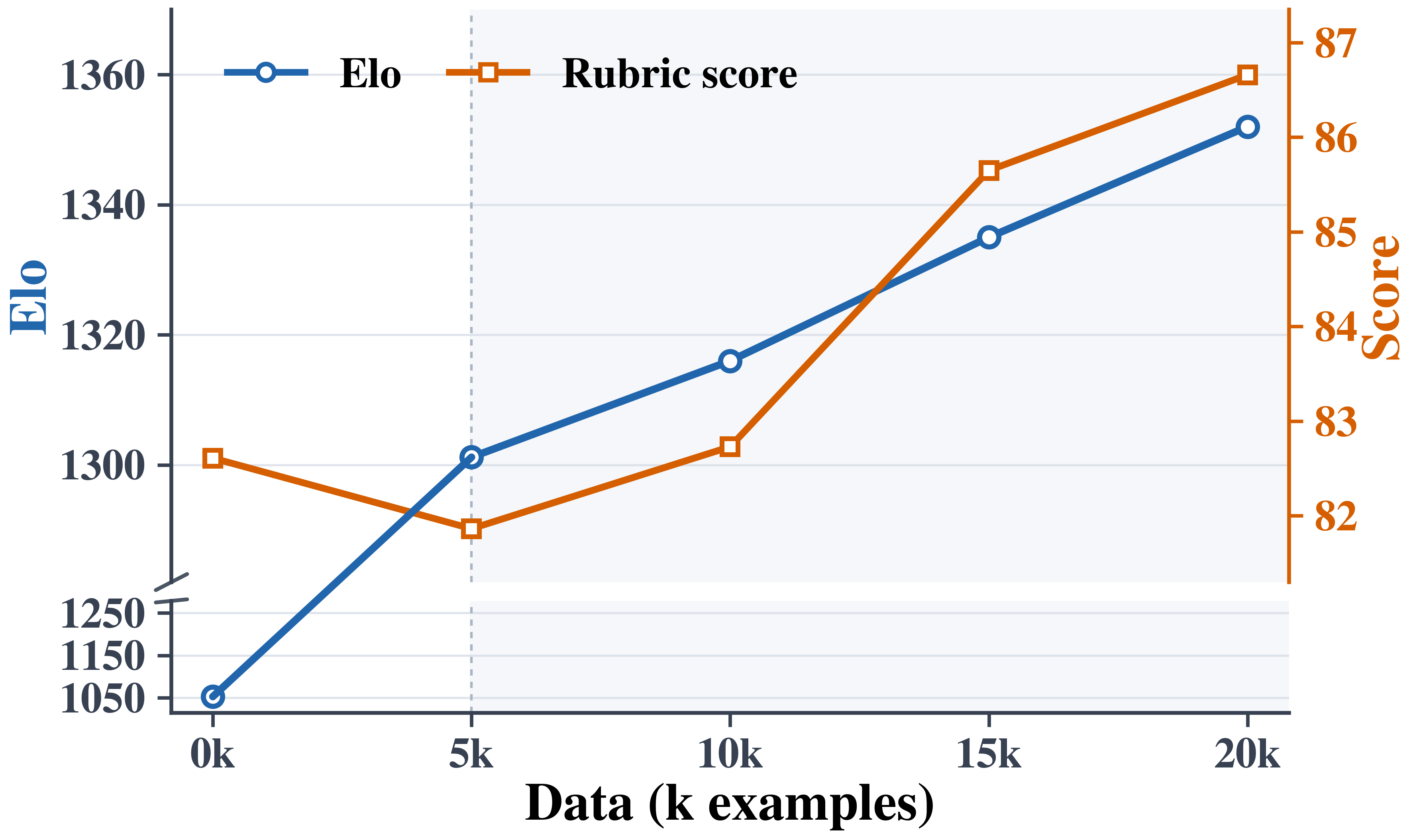}

\caption{\textbf{Performance of Fx-Work-35B trained with varying amounts of data on GDPvalAA-v2.} As the amount of training data increases, Fx-Work-35B consistently improves on GDPvalAA-v2, demonstrating that WorkGenesis-synthesized data exhibit favorable scaling behavior.}
\label{fig:scaling}
\end{minipage}

\end{figure}

\textbf{Fx-Work-35B achieves a higher average score at a lower per-task cost.} Figure~\ref{fig:cost} shows the average cost of completing one occupational task and the average score on the three benchmarks for all models in Table~\ref{tab:main}. Fx-Work-35B achieves an average score of 31.00 at a cost of only \$0.77. Compared to models with similar scores (e.g. DeepSeek-V4-Pro-Preview), Fx-Work-35B reduces the per-task cost by \$0.44 (36\% lower), while compared to models with similar per-task costs (e.g. Nex-N2-mini), Fx-Work-35B improves overall performance on occupational tasks by 13.26 points (74\% higher). Overall, after training on occupational tasks synthesized by WorkGenesis, Fx-Work-35B is a highly cost-effective model capable of completing daily occupational work, further demonstrating that the occupational tasks synthesized by WorkGenesis are close to real-world tasks and of high quality.

\begin{figure}[t]
\centering

    \includegraphics[width=1\linewidth]{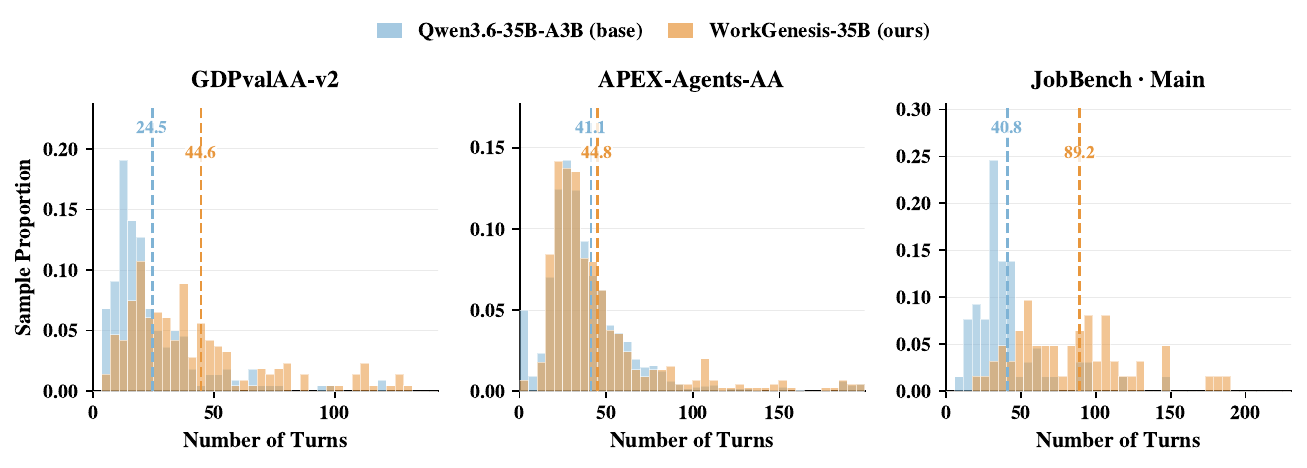}

\caption{\textbf{Number of turns per task before and after training.} The horizontal axis is the number of turns required to complete a task and the vertical axis is the share of tasks. Fx-Work-35B exhibits stronger long-horizon task capabilities.}
\label{fig:turns}

\end{figure}

\textbf{Fx-Work-35B exhibits stronger long-horizon task capabilities.} Figure~\ref{fig:turns} compares the distribution of the number of turns used by different models to complete an occupational task. Before training, Qwen3.6-35B-A3B uses an average of 24.5, 41.1, 40.8 turns to complete a task on three benchmarks respectively. After training on occupational scenarios synthesized by WorkGenesis, Fx-Work-35B, when completing the same tasks, increases its average number of turns to 44.6, 44.8, 89.2 respectively, and the variation across different tasks is relatively large, while the overall turn distribution remains relatively balanced. This indicates that: (1) Fx-Work-35B invests more thinking into the details of completing occupational tasks and performs more verification, all of which contribute to the improvement in the quality of its task completion; and (2) Fx-Work-35B does not blindly increase the number of turns or perform verification mindlessly. Instead, after training on a large number of occupational tasks synthesized by WorkGenesis that closely match real-world scenarios, it can solve simple occupational tasks more quickly, while also solving complex long-horizon occupational tasks better and more meticulously.

\subsection{Discussion and Analysis}
\label{sec:discussion}

\textbf{WorkGenesis-synthesized data show improving performance as data volume increases.} Figure~\ref{fig:scaling} shows the effect of increasing the number of WorkGenesis-synthesized occupational scenarios, and thereby the training data volume, under identical training hyperparameters. We train Fx-Work-35B with 5K, 10K, 15K, and 20K WorkGenesis-synthesized occupational scenarios, and report its rubric and Elo scores on GDPvalAA-v2. As the amount of data increases, the Elo score of Fx-Work-35B improves steadily. For the rubric score, we observe a slight drop at 5K, followed by a consistent upward trend. We attribute this initial drop to the model's capabilities not yet being fully developed, such that some tasks cannot be completed within the prescribed turn budget, which reduces the rubric score that emphasizes task completion. In contrast, Elo focuses more on the quality of each completed task and is therefore less affected. As the data volume continues to increase and the model's capabilities improve, the rubric score also exhibits a steadily improving trend.

\begin{figure}[t]
\centering
\begin{minipage}[t]{0.48\linewidth}
\centering
\includegraphics[width=1\linewidth]{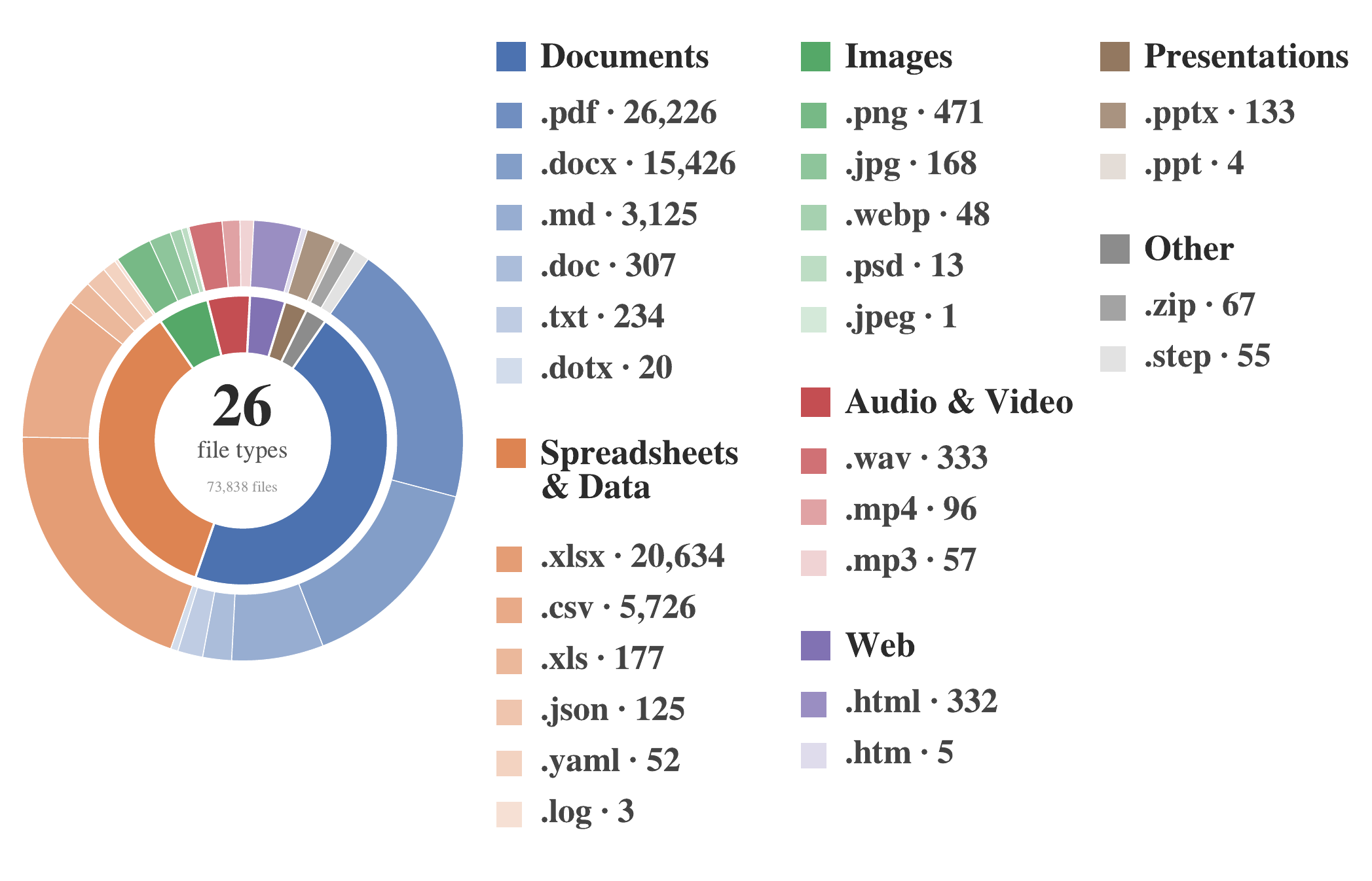}

\caption{\textbf{Diversity Analysis of Occupational Tasks Synthesized by WorkGenesis.} The occupational tasks synthesized by WorkGenesis are grounded in the characteristics of different occupations and use a diverse set of files as task reference files.}
\label{fig:diversity}
\end{minipage}\hfill
\begin{minipage}[t]{0.48\linewidth}
\centering
    \includegraphics[width=1\linewidth]{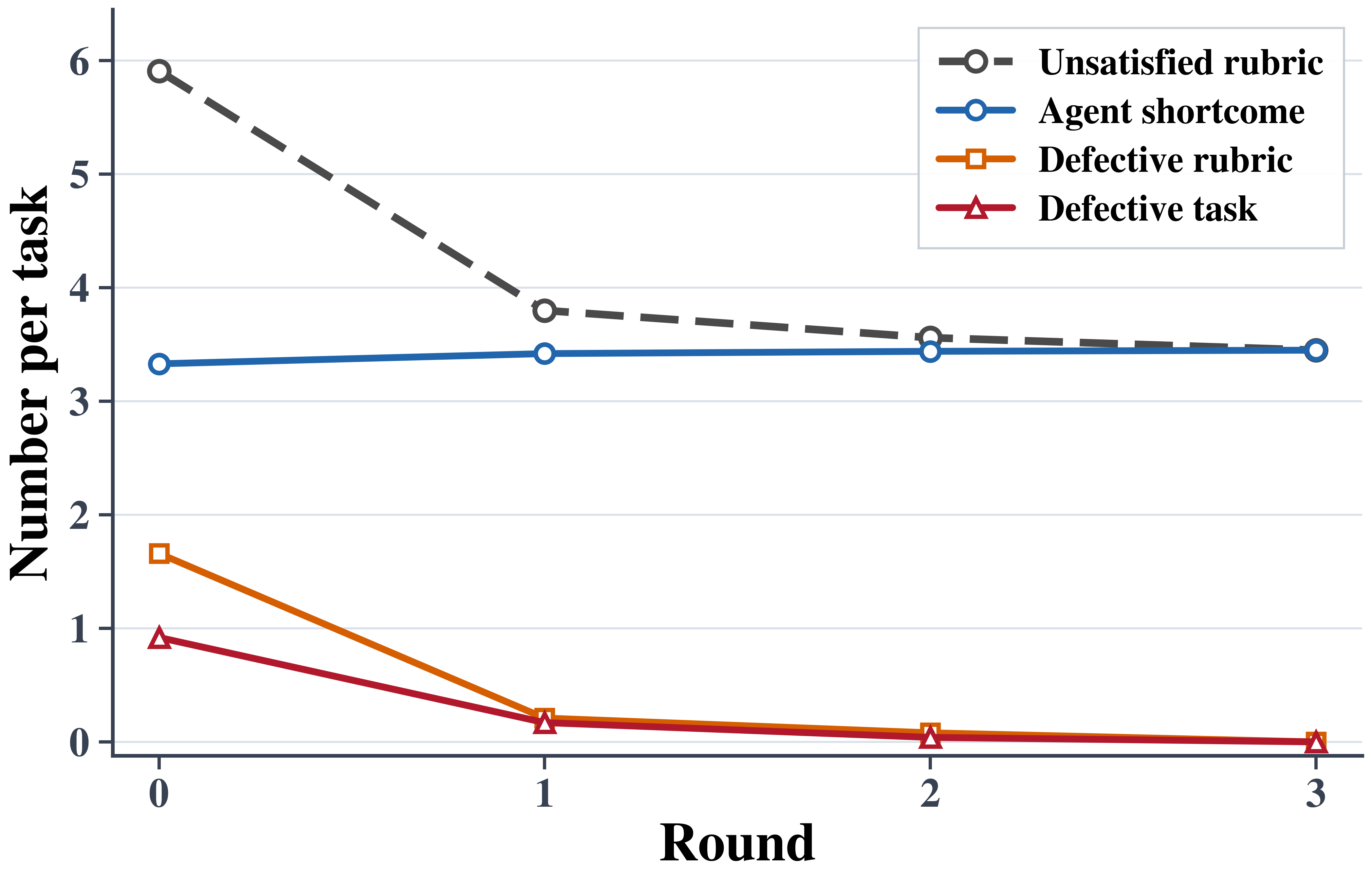}
 
\caption{\textbf{Effect of Execution-Guided Consistency Verification.} The horizontal axis is the round of verification and the vertical axis is the average number of rubric items per unit of work attributed to $\mathsf{Rubric}$ or $\mathsf{Task}$. The count decreases across rounds. The average number of rubric items is 42.10 per task.}
\label{fig:repair}
\end{minipage}

\end{figure}

\textbf{WorkGenesis synthesizes diverse occupational work scenarios.} Figure~\ref{fig:diversity} shows the distribution of reference files used across all occupational tasks. As shown in Figure~\ref{fig:diversity}, the occupational tasks synthesized by WorkGenesis mainly use PDFs, spreadsheets, and Word documents as attachments, but also include many professional reference files such as $\texttt{.step}$ and $\texttt{.psd}$ used in specialized occupational scenarios, demonstrating the diversity of occupational tasks.

\textbf{Consistency verification improves scenario quality across rounds.} Figure~\ref{fig:repair} tracks the average number of rubric items per unit of work that are attributed to $\mathsf{Agent}$, $\mathsf{Rubric}$, and $\mathsf{Task}$. The defect count ($\mathsf{Rubric}$ and $\mathsf{Task}$) falls from 1.66 and 0.92 after the initial version to zero after three rounds of verification. The decline shows that the verification step removes the defects it targets, and the stopping rule of the pipeline keeps every accepted unit of work at zero defective items, demonstrating the effectiveness of the execution-guided consistency verification.

\section{Conclusion}
\label{sec:conclusion}

We introduced WorkGenesis, a framework that constructs executable occupational work from real-world artifacts. Evidence-Based Work Construction uses O*NET occupations to retrieve public anchor files and builds the request, companion materials, and itemized rubric around them, while Execution-Guided Consistency Verification repairs task and rubric defects exposed by a reference execution until the work is internally consistent. From 65 occupations, WorkGenesis produces 20K units of work whose trajectories train Fx-Work-35B. The model raises GDPval Elo from 1053 to 1352, APEX-Agents pass@1 from 13.4\% to 25.2\%, and JobBench Main/Easy from 14.99/56.02 to 25.19/70.06, leading all comparable-scale baselines on every reported metric and surpassing the 1.6T DeepSeek-V4-Pro-Preview on four of the five. These results establish evidence-grounded, executable work as a practical source of training data for working agents, and we view extending construction to more occupations and to reinforcement learning as promising next steps.

\clearpage

\subsection*{AI use statement}

In this work, we used generative AI tools (LLM) to construct occupational work
scenarios, synthesize companion files, formulate task requests and evaluation
rubrics, execute tasks, evaluate deliverables, identify inconsistencies, and
generate training trajectories. 
All AI-assisted outputs were reviewed through execution-based verification,
item-level rubric checking, manual inspection, and reproducible evaluation
scripts. The authors take responsibility for the final content of this
paper, including all claims, code, data, and artifacts produced with the aid
of generative AI.

\clearpage

\bibliography{references}

@misc{openai2025gdpval,
  author = {{OpenAI}},
  title = {Measuring the Performance of Our Models on Real-World Tasks},
  year = {2025},
  month = sep,
  howpublished = {\url{https://openai.com/index/gdpval/}}
}

@article{jobbench2026,
  title={JobBench: Aligning Agent Work With Human Will},
  author={Li, Yuetai and Feng, Yichen and Xu, Zhangchen and Ma, Zixian and Zheng, Kaiyuan and Jiang, Fengqing and Sun, Xinghua and Shao, Rulin and Chen, Zichen and Huang, Yue and others},
  journal={arXiv preprint arXiv:2605.26329},
  year={2026}
}

@article{zuo2026qwenagentworld,
  title={Qwen-AgentWorld: Language World Models for General Agents},
  author={Zuo, Yuxin and Xiao, Zikai and Sheng, Li and Huang, Fei and Tu, Jianhong and Liu, Yuxuan and Tang, Tianyi and Hu, Xiaomeng and Su, Yang and Lan, Qingfeng and others},
  journal={arXiv preprint arXiv:2606.24597},
  year={2026}
}

@misc{onetdatabase,
  author = {{O*NET Resource Center}},
  title = {The {O*NET} Database},
  year = {n.d.},
  howpublished = {\url{https://www.onetcenter.org/database.html}}
}

@inproceedings{drouin2024workarena,
  title={WorkArena: How capable are web agents at solving common knowledge work tasks?},
  author={Drouin, Alexandre and Gasse, Maxime and Caccia, Massimo and Laradji, Issam H and Del Verme, Manuel and Marty, Tom and Boisvert, L{\'e}o and Thakkar, Megh and Cappart, Quentin and Vazquez, David and others},
  journal={arXiv preprint arXiv:2403.07718},
  year={2024}
}

@inproceedings{xie2024osworld,
  title={Osworld: Benchmarking multimodal agents for open-ended tasks in real computer environments},
  author={Xie, Tianbao and Zhang, Danyang and Chen, Jixuan and Li, Xiaochuan and Zhao, Siheng and Cao, Ruisheng and Hua, Toh J and Cheng, Zhoujun and Shin, Dongchan and Lei, Fangyu and others},
  journal={Advances in Neural Information Processing Systems},
  volume={37},
  pages={52040--52094},
  year={2024}
}

@inproceedings{wang2023selfinstruct,
  title={Self-instruct: Aligning language models with self-generated instructions},
  author={Wang, Yizhong and Kordi, Yeganeh and Mishra, Swaroop and Liu, Alisa and Smith, Noah A and Khashabi, Daniel and Hajishirzi, Hannaneh},
  booktitle={Proceedings of the 61st annual meeting of the association for computational linguistics (volume 1: long papers)},
  pages={13484--13508},
  year={2023}
}

@inproceedings{chen2026dreamgym,
  title={Scaling agent learning via experience synthesis},
  author={Chen, Zhaorun and Zhao, Zhuokai and Zhang, Kai and Liu, Bo and Qi, Qi and Wu, Yifan and Kalluri, Tarun and Cao, Xuefei and Xiong, Yuanhao and Tong, Haibo and others},
  booktitle={International Conference on Learning Representations},
  volume={2026},
  pages={121394--121420},
  year={2026}
}

@article{madaan2023selfrefine,
  title={Self-refine: Iterative refinement with self-feedback},
  author={Madaan, Aman and Tandon, Niket and Gupta, Prakhar and Hallinan, Skyler and Gao, Luyu and Wiegreffe, Sarah and Alon, Uri and Dziri, Nouha and Prabhumoye, Shrimai and Yang, Yiming and others},
  journal={Advances in neural information processing systems},
  volume={36},
  pages={46534--46594},
  year={2023}
}

@article{shinn2023reflexion,
  title={Reflexion: Language agents with verbal reinforcement learning},
  author={Shinn, Noah and Cassano, Federico and Gopinath, Ashwin and Narasimhan, Karthik and Yao, Shunyu},
  journal={Advances in neural information processing systems},
  volume={36},
  pages={8634--8652},
  year={2023}
}

@article{pandya2026echoverse,
  title={Echoverse: Deep, Evolving Environments for Training Computer-Use Agents at Scale},
  author={Pandya, Yash and Gupta, Sahil and Harne, Sarthak and Yadav, Archana and Chourasia, Kavyansh and Mozannar, Hussein and Vineet, Vibhav and Abdali, Sara and Rosset, Corby and Lara, Yash and others},
  journal={arXiv preprint arXiv:2607.28074},
  year={2026}
}

@article{xu2024wizardlm,
  title={Wizardlm: Empowering large language models to follow complex instructions},
  author={Xu, Can and Sun, Qingfeng and Zheng, Kai and Geng, Xiubo and Zhao, Pu and Feng, Jiazhan and Tao, Chongyang and Jiang, Daxin},
  journal={arXiv e-prints},
  pages={arXiv--2304},
  year={2023}
}

@article{chan2024personahub,
  title={Scaling synthetic data creation with 1,000,000,000 personas},
  author={Ge, Tao and Chan, Xin and Wang, Xiaoyang and Yu, Dian and Mi, Haitao and Yu, Dong},
  journal={arXiv preprint arXiv:2406.20094},
  year={2024}
}

@inproceedings{xu2025magpie,
  title={Magpie: Alignment data synthesis from scratch by prompting aligned llms with nothing},
  author={Xu, Zhangchen and Jiang, Fengqing and Niu, Luyao and Deng, Yuntian and Poovendran, Radha and Choi, Yejin and Lin, Bill Yuchen},
  booktitle={International Conference on Learning Representations},
  volume={2025},
  pages={76346--76382},
  year={2025}
}

@article{xu2025theagentcompany,
  title={Theagentcompany: benchmarking llm agents on consequential real world tasks},
  author={Xu, Frank Fangzheng and Song, Yufan and Li, Boxuan and Tang, Yuxuan and Jain, Kritanjali and Bao, Mengxue and Wang, Zora and Zhou, Xuhui and Guo, Zhitong and Cao, Murong and others},
  journal={Advances in Neural Information Processing Systems},
  volume={38},
  year={2026}
}

@inproceedings{huang2025crmarena,
  title={Crmarena: Understanding the capacity of llm agents to perform professional crm tasks in realistic environments},
  author={Huang, Kung-Hsiang and Prabhakar, Akshara and Dhawan, Sidharth and Mao, Yixin and Wang, Huan and Savarese, Silvio and Xiong, Caiming and Laban, Philippe and Wu, Chien-Sheng},
  booktitle={Proceedings of the 2025 Conference of the Nations of the Americas Chapter of the Association for Computational Linguistics: Human Language Technologies (Volume 1: Long Papers)},
  pages={3830--3850},
  year={2025}
}

@article{yao2024taubench,
  title={$tau $-bench: A Benchmark for Tool-Agent-User Interaction in Real-World Domains},
  author={Yao, Shunyu and Shinn, Noah and Razavi, Pedram and Narasimhan, Karthik},
  journal={arXiv preprint arXiv:2406.12045},
  year={2024}
}

@inproceedings{patwardhan2025gdpval,
  title={Gdpval: Evaluating ai model performance on real-world economically valuable tasks},
  author={Patwardhan, Tejal and Dias, Rachel and Proehl, Elizabeth and Kim, Grace and Wang, Michele and Watkins, Olivia and Fishman, Simon and Aljubeh, Marwan and Thacker, Phoebe and Fauconnet, Laurance and others},
  booktitle={International Conference on Learning Representations},
  volume={2026},
  pages={24005--24040},
  year={2026}
}

@article{vidgen2026apex,
  title={APEX-agents},
  author={Vidgen, Bertie and Mann, Austin and Fennelly, Abby and Stanly, John Wright and Rothman, Lucas and Burstein, Marco and Benchek, Julien and Ostrofsky, David and Ravichandran, Anirudh and Sur, Debnil and others},
  journal={arXiv preprint arXiv:2601.14242},
  year={2026}
}

@misc{qwen36_35b_a3b,
    title = {{Qwen3.6-35B-A3B}: Agentic Coding Power, Now Open to All},
    url = {https://qwen.ai/blog?id=qwen3.6-35b-a3b},
    author = {{Qwen Team}},
    month = {April},
    year = {2026}
}

@misc{qwen38_max,
    title = {{Qwen3.8-Max}: A New Bar for Coding and Cowork},
    url = {https://qwen.ai/blog?id=qwen3.8},
    author = {{Qwen Team}},
    month = {August},
    year = {2026}
}

@misc{zai2026glm52,
  title={{GLM-5.2}: Built for Long-Horizon Tasks},
  author={{Z.ai}},
  year={2026},
  month={June},
  url={https://z.ai/blog/glm-5.2}
}

@misc{deepseek2026v4,
  title={DeepSeek V4 Preview Release},
  author={{DeepSeek Team}},
  year={2026},
  month={April},
  url={https://api-docs.deepseek.com/news/news260424/}
}

@misc{openai2026gpt55,
  title={{GPT-5.5} System Card},
  author={{OpenAI}},
  year={2026},
  month={April},
  url={https://deploymentsafety.openai.com/gpt-5-5/gpt-5-5.pdf}
}

@article{team2026kimi,
  title={Kimi k3: Open frontier intelligence},
  author={Team, Kimi and Bai, Tongtong and Bai, Yifan and Bao, Yiping and Cai, Jianfeng and Cai, Xinyuan and Cao, Peizhou and Cao, Yuxuan and Chai, Ziwei and Charles, Y and others},
  journal={arXiv preprint arXiv:2607.24653},
  year={2026}
}

@article{bai2026agentsa1,
  title={Scaling the Horizon, Not the Parameters: Reaching Trillion-Parameter Performance with a 35B Agent},
  author={Bai, Lei and Cao, Zongsheng and Chen, Yang and Cui, Zhiyao and Du, Shangheng and Fan, Yue and Feng, Shiyang and Guo, Zijie and He, Haonan and He, Liang and others},
  journal={arXiv preprint arXiv:2606.30616},
  year={2026}
}

@misc{nexagi2026nexn2,
  title={Nex-N2: Thinking, Built for Action},
  author={{Nex AGI}},
  year={2026},
  month={June},
  url={https://nex-agi.com/}
}

@misc{apodex2026,
  title={Apodex-1.0: A Verification-Centric Agent Team for Discoverative Intelligence},
  author={{Apodex Team}},
  year={2026},
  month={June},
  url={https://www.apodex.com/blog/apodex-1.0}
}

@article{chen2026occamy10openparetofrontier35b,
  title={Occamy-1.0: Open Pareto-frontier 35B Intelligence for Co-work},
  author={Chen, Wenhui and Cheng, Shiwen and Dong, Hao and Duan, Chenda and Feng, Ruixiang and Guan, Zhong and Guo, Boqiang and Han, Xueyuan and Hao, Haojie and Huang, Liangmeng and others},
  journal={arXiv preprint arXiv:2609.11977},
  year={2026},
  url = {https://arxiv.org/abs/2609.11977}
}

@article{scicodeverified,
  title={SciCode-Verified: How Benchmark Defects Underestimated the Scientific-Coding Ability of Language Models},
  author={Hu, Sihan and Huang, Lyuhan and Deng, Youjin and Chen, Kun},
  journal={arXiv preprint arXiv:2608.04975},
  year={2026},
  url={https://arxiv.org/abs/2608.04975}
}

@misc{artificialanalysis2026gdpvalaa,
  title={{GDPval} Agentic Evaluation and the {Stirrup} Agent Harness},
  author={{Artificial Analysis}},
  howpublished={\url{https://artificialanalysis.ai/}},
  year={2026}
}
\bibliographystyle{plainnat}

\clearpage

\appendix
We describe the implementation details of WorkGenesis in detail in~\ref{app:impl_details}, provide further analysis of the occupational tasks synthesized by WorkGenesis in~\ref{app:analysis}, present two case studies of occupational tasks synthesized by WorkGenesis in~\ref{app:task-overview}, and show the prompts used by each component of WorkGenesis in~\ref{app:prompts}.

\section{Implementation Details}
\label{app:impl_details}
In this section, we provide a detailed description of the synthesis details of WorkGenesis, the training details of Fx-Work-35B, and the evaluation details.

\subsection{Synthesis Details}
Throughout the process of synthesizing occupational tasks with WorkGenesis, we consistently use GLM-5.2 as the LLM driving each module. We use occupational information from 65 O*NET occupations as the occupation playbook for the Synthesis request stage, and fix the tools to web\_search, web\_fetch, code\_exec, and view\_image. Table~\ref{tab:occupation-distribution} provides a detailed list of the 65 O*NET occupations we use. We use the Serp API\footnote{https://www.serpapi.com} as the underlying wrapper for the web\_search and web\_fetch tools. When rendering the deliverable, we use stirrup as the agent harness and GLM-5.2 as the backend LLM. For the LLM prompts of each WorkGenesis module, please refer to Section~\ref{app:prompts}.


\begin{table*}[t]
\centering
\caption{The 65 occupation categories used in WorkGenesis.
Sectors: FIN = Finance and Insurance,
GOV = Government, HCS = Health Care and Social Assistance,
INF = Information, MFG = Manufacturing,
PST = Professional, Scientific, and Technical Services,
RER = Real Estate and Rental and Leasing,
RET = Retail Trade, WHL = Wholesale Trade,
EDU = Educational Services,
MIN = Mining, Quarrying, and Oil and Gas Extraction.}
\label{tab:occupation-distribution}
\small
\setlength{\tabcolsep}{4pt}
\renewcommand{\arraystretch}{1.15}

\begin{tabularx}{\textwidth}{
@{}l >{\RaggedRight\arraybackslash}X
@{\hspace{2em}}
l >{\RaggedRight\arraybackslash}X@{}}

\toprule
Sector & Occupation & Sector & Occupation \\
\midrule

FIN & Customer Service Representatives
& FIN & Financial Managers \\

FIN & Financial and Investment Analysts
& FIN & Personal Financial Advisors \\

FIN & Securities, Commodities, and Financial Services Sales Agents
& GOV & Administrative Services Managers \\

GOV & Child, Family, and School Social Workers
& GOV & Compliance Officers \\

GOV & Court, Municipal, and License Clerks
& GOV & First-Line Supervisors of Police and Detectives \\

GOV & Public Safety Telecommunicators
& GOV & Recreation Workers \\

HCS & First-Line Supervisors of Office and Administrative Support Workers
& HCS & Medical Secretaries and Administrative Assistants \\

HCS & Medical and Health Services Managers
& HCS & Nurse Practitioners \\

HCS & Registered Nurses
& INF & Audio and Video Technicians \\

INF & Editors
& INF & Film and Video Editors \\

INF & News Analysts, Reporters, and Journalists
& INF & Producers and Directors \\

INF & Web Administrators
& MFG & Buyers and Purchasing Agents \\

MFG & First-Line Supervisors of Production and Operating Workers
& MFG & Industrial Engineers \\

MFG & Mechanical Engineering Technologists and Technicians
& MFG & Mechanical Engineers \\

MFG & Shipping, Receiving, and Inventory Clerks
& PST & Accountants and Auditors \\

PST & Biostatisticians
& PST & Bookkeeping, Accounting, and Auditing Clerks \\

PST & Civil Engineers
& PST & Computer and Information Research Scientists \\

PST & Computer and Information Systems Managers
& PST & Computer User Support Specialists \\

PST & Data Entry Keyers
& PST & Human Resources Specialists \\

PST & Lawyers
& PST & Management Analysts \\

PST & Project Management Specialists
& PST & Secretaries and Administrative Assistants,
Except Legal, Medical, and Executive \\

PST & Social Science Research Assistants
& PST & Software Developers \\

PST & Statisticians
& PST & Technical Writers \\

PST & Training and Development Specialists
& RER & Concierges \\

RER & Counter and Rental Clerks
& RER & Property, Real Estate, and Community Association Managers \\

RER & Real Estate Brokers
& RER & Real Estate Sales Agents \\

RET & First-Line Supervisors of Retail Sales Workers
& RET & General and Operations Managers \\

RET & Online Merchants
& RET & Pharmacists \\

RET & Private Detectives and Investigators
& WHL & First-Line Supervisors of Non-Retail Sales Workers \\

WHL & Order Clerks
& WHL & Sales Managers \\

WHL & Sales Representatives, Wholesale and Manufacturing,
Except Technical and Scientific Products
& WHL & Sales Representatives, Wholesale and Manufacturing,
Technical and Scientific Products \\

WHL & Supply Chain Managers
& EDU & Sociology Teachers, Postsecondary \\

MIN & Petroleum Engineers
& & \\

\bottomrule
\end{tabularx}
\end{table*}

\subsection{Training Details}
\begin{figure}[t]
\centering
    \includegraphics[width=1\linewidth]{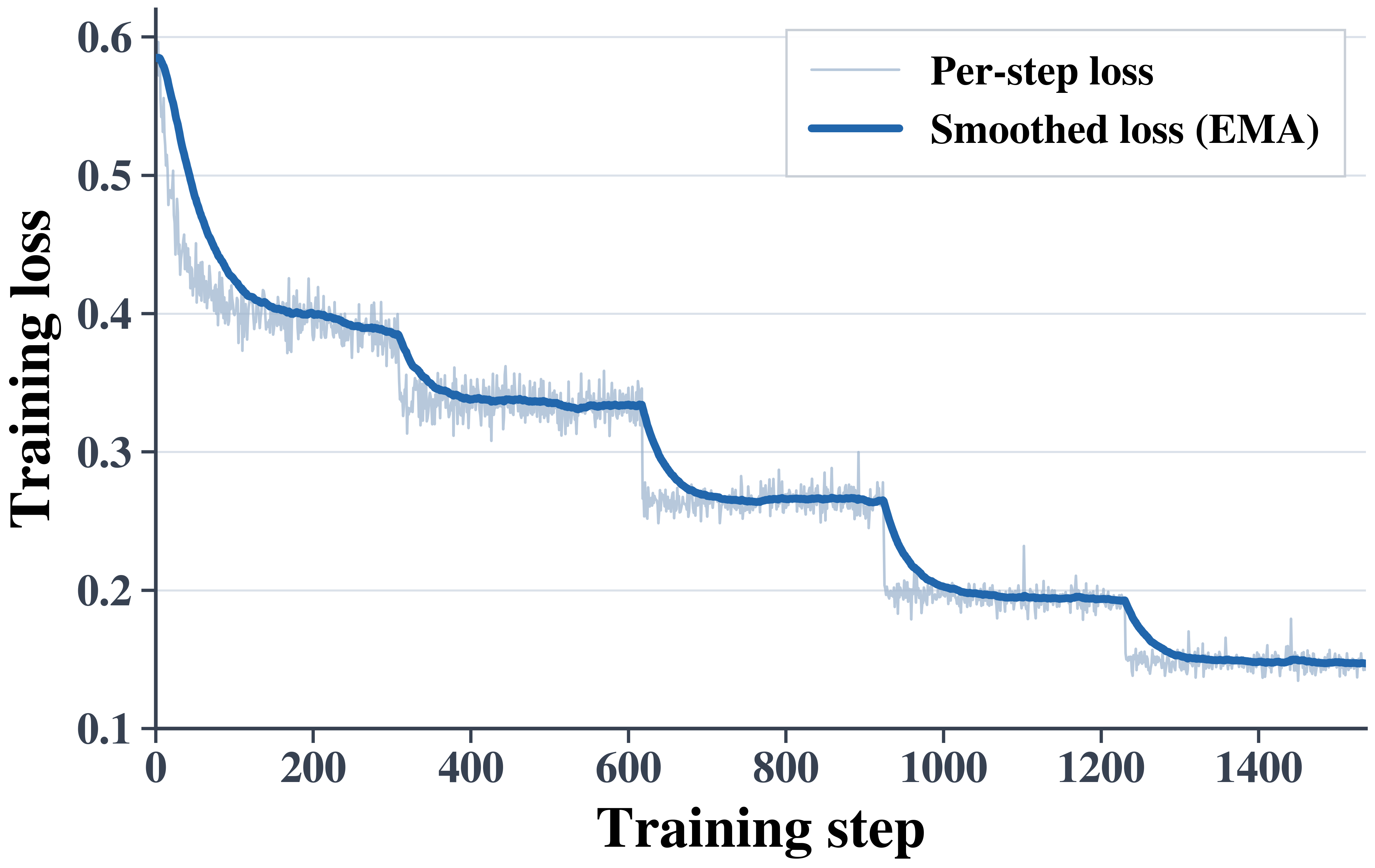}
\caption{\textbf{Training loss curve of Fx-Work-35B.}}
\label{fig:training_loss}
\end{figure}

We train Qwen3.6-35B-A3B using standard supervised fine-tuning (SFT). The training loss curve is shown in Figure~\ref{fig:training_loss}, and the training hyperparameters are summarized in Table~\ref{tab:training-hyperparams}.

\begin{table}[t]
  \centering
  \small
  \setlength{\tabcolsep}{6pt}
  \renewcommand{\arraystretch}{1.05}
  \caption{Key training hyperparameters for training Fx-Work-35B.}
  \label{tab:training-hyperparams}
  \begin{tabular}{@{}ll@{}}
    \toprule
    \multicolumn{2}{@{}l}{\textit{Model and initialization}} \\
    \midrule
    Base model               & Qwen3.6-35B-A3B (sparse MoE) \\
    \midrule
    \multicolumn{2}{@{}l}{\textit{Training data and mixture}} \\
    \midrule
    Update steps             & 1{,}537 \\
    Packing bin              & 256{,}000 tokens (16k per GPU over CP) \\
    Sampling                 & shuffled and length-balanced per rollout \\
    \midrule
    \multicolumn{2}{@{}l}{\textit{Optimization}} \\
    \midrule
    Optimizer                & Adam, $\beta_1 = 0.9$, $\beta_2 = 0.98$ \\
    Learning rate            & $5\times10^{-5}$, cosine decay to $10^{-6}$ \\
    Warmup                   & 10\% of steps (linear) \\
    Weight decay             & 0.1 \\
    Optimizer state          & distributed, CPU offload, precision-aware \\
    \midrule
    \multicolumn{2}{@{}l}{\textit{Batching, parallelism and infrastructure}} \\
    \midrule
    Rollout / global batch   & 64 / 64 (1 sample per prompt) \\
    Parallelism              & TP 1 $\times$ PP 2 $\times$ CP 16, EP 8, ET 1 \\
    Sequence parallelism     & enabled; dynamic batching, 16k tokens/GPU \\
    Hardware                 & 4 nodes $\times$ 8 H100 GPUs $=$ 32 GPUs \\
    Activation recompute     & full, uniform, 1 layer \\
    \bottomrule
  \end{tabular}
\end{table}

\subsection{Evaluation Details}
\textbf{GDPvalAA-v2.} When evaluating GDPvalAA-v2, we strictly follow the official evaluation criteria published by Artificial Analysis\footnote{https://artificialanalysis.ai/methodology/intelligence-benchmarking}. Specifically, we use the official harness Stirrup provided by Artificial Analysis\footnote{https://github.com/ArtificialAnalysis/Stirrup}, adopt a pre-built image with a large number of packages installed as the startup image for the E2B sandbox, and align details such as tools and context compression parameters. For deliverable scoring, we use GPT-5.5 and Gemini-3.6-flash as scoring models, respectively, and score each deliverable based on the task prompt, parsed task attachments, parsed deliverable attachments, and the rubric, then take the average as the final score. For ELO computation, for models already evaluated by Artificial Analysis, we directly use their scores from the Artificial Analysis snapshot on August 4, 2026; for models not yet evaluated, we compute ELO scores based on the snapshot, our evaluation results on 15 commonly used models, and Artificial Analysis's ELO computation model. When computing weighted scores, ELO scores are normalized using the official formula, \((\mathrm{Elo} - 500) / 2000 \times 100\%\).

\textbf{APEX-Agents-AA.} When evaluating APEX-Agents-AA, we also follow the official Artificial Analysis configuration and use the same Stirrup harness and sandbox setup described above. The benchmark contains 480 professional tasks; following Artificial Analysis, we exclude the two Investment Banking worlds that rely on real-time external data feeds (World 244 and World 246), resulting in 452 tasks spanning Law (160), Management Consulting (160), and Investment Banking (132). Each task is paired with a rubric consisting of binary criteria. For every attempt, an LLM judge (Gemini-3.6-Flash) grades each criterion as 0 or 1 based on the task prompt, the task attachments, and the parsed deliverable, and an attempt is counted as a pass only if all criteria are satisfied. We run 3 independent repeats per task and report pass@1, computed as the fraction of passing attempts over all $452 \times 3$ attempts.

\textbf{JobBench.} When evaluating JobBench, we also use Stirrup as the agent harness. For deliverable evaluation, we use GPT-5.5 and Gemini-3.6-flash as scoring models, and score each deliverable based on the task prompt, parsed task attachments, parsed deliverable attachments, and the rubric, then take the average as the final score.

\textbf{Aggregation Score.} To compute the aggregation score, we use the average of the normalized ELO score on GDPvalAA-v2, the pass rate on APEX-Agents-AA, and the score on the JobBench main split as the Aggregation Score across the three benchmarks.

\section{Analysis and results}
\label{app:analysis}
\begin{figure*}[t]
\centering
\centering
    \includegraphics[width=1\linewidth]{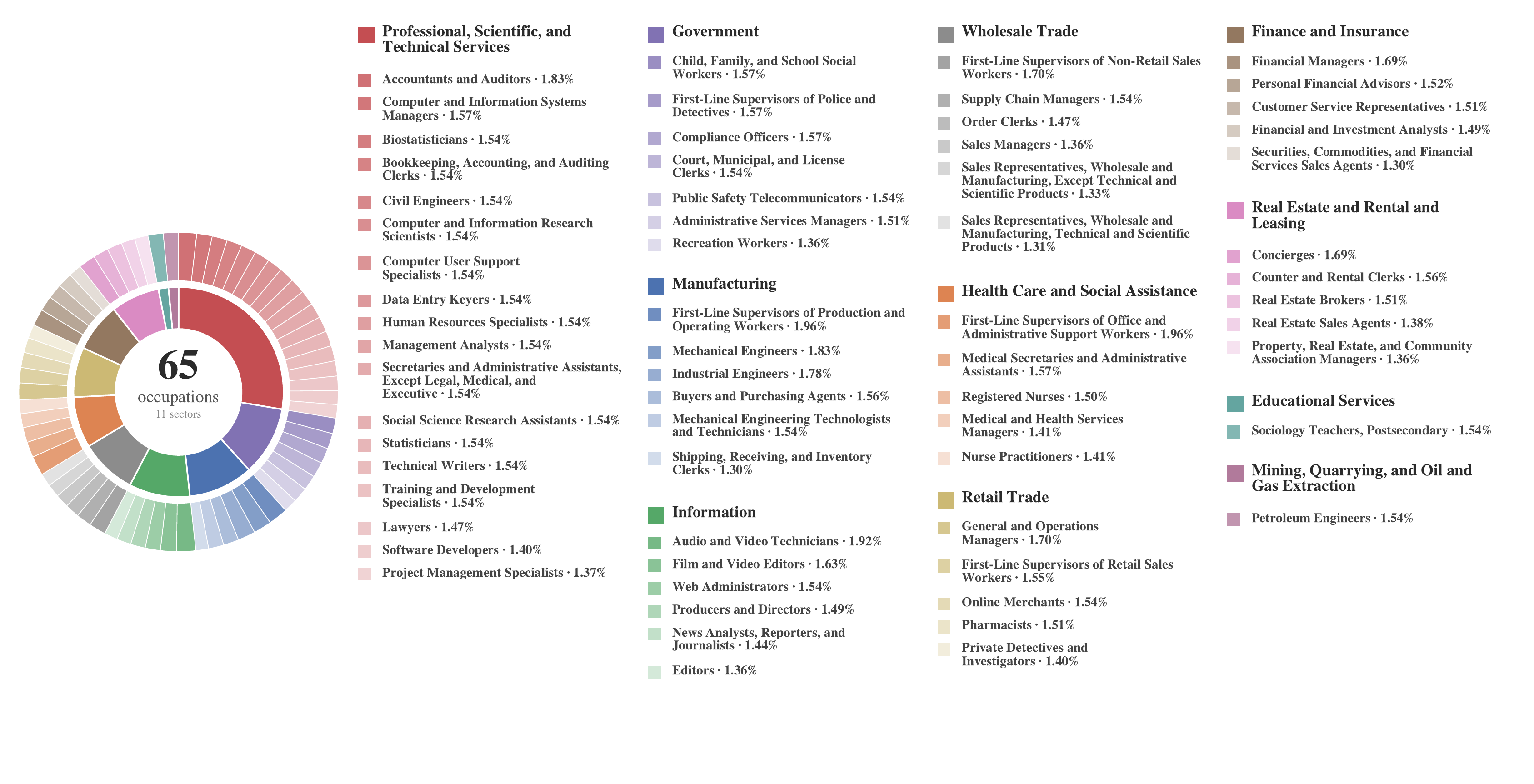}
\caption{\textbf{Diversity of the constructed work.} (a) Distribution of the 20K units of work over the 65 occupations.}
\label{fig:diversity2}
\end{figure*}

In this section, we show more analysis about the work scenario generated by WorkGenesis. In Figure~\ref{fig:diversity2}, we show 
the distribution of the 20K occupational tasks synthesized by WorkGenesis. The occupational tasks synthesized by WorkGenesis are evenly distributed, covering 65 occupations.

%
%
%

\definecolor{tasknavy}{HTML}{112240}
\definecolor{taskink}{HTML}{243A55}
\definecolor{accent}{HTML}{175CD3}
\definecolor{band}{HTML}{F2F4F7}
\definecolor{prompttint}{HTML}{F7F9FC}
\definecolor{evidencetint}{HTML}{F3F8F6}
\definecolor{rubrictint}{HTML}{FBF8F1}
\definecolor{penaltytint}{HTML}{FDF0EE}
\definecolor{penalty}{HTML}{B42318}
\definecolor{muted}{HTML}{667085}
\definecolor{themebg}{HTML}{EAEFF5}

\newtcolorbox{TaskMeta}[1][]{enhanced,breakable,boxrule=.5pt,arc=1.2pt,
  colback=band,colframe=tasknavy,coltitle=white,colbacktitle=tasknavy,
  fonttitle=\bfseries\small,left=6pt,right=6pt,top=5pt,bottom=6pt,title=#1}
\newtcolorbox{PromptBox}[1][]{enhanced,breakable,boxrule=.5pt,arc=1.2pt,
  colback=prompttint,colframe=accent,coltitle=taskink,colbacktitle=prompttint,
  fonttitle=\bfseries\small,left=6pt,right=6pt,top=5pt,bottom=6pt,title=#1}
\newtcolorbox{EvidenceBox}[1][]{enhanced,breakable,boxrule=.5pt,arc=1.2pt,
  colback=evidencetint,colframe=taskink,coltitle=taskink,colbacktitle=evidencetint,
  fonttitle=\bfseries\small,left=6pt,right=6pt,top=5pt,bottom=6pt,title=#1}
\newtcolorbox{RubricBox}[1][]{enhanced,breakable,boxrule=.5pt,arc=1.2pt,
  colback=rubrictint,colframe=taskink,coltitle=taskink,colbacktitle=rubrictint,
  fonttitle=\bfseries\small,left=6pt,right=6pt,top=5pt,bottom=6pt,title=#1}
\newcommand{\evidencecaption}[1]{\par\smallskip\centering\footnotesize\color{muted}#1\par}

\section{Case Study}
\label{app:task-overview}
In this section, we present in detail two specific occupational tasks synthesized by WorkGenesis.

\subsection{Task A $\cdot$ Line 4 roll-former guard retrofit review}
\label{app:task-a}

\begin{TaskMeta}[Task A $\cdot$ file-backed mechanical-engineering review]
\begin{tabularx}{\linewidth}{@{}>{\bfseries}p{.205\linewidth}X@{}}
Occupation & Mechanical Engineers\\
Work request & Line 4 roll-former guard retrofit decision review (email from the manufacturing-engineering manager)\\
Task identifier & \texttt{3ef2d8a9-59d6-4d32-8a1f-fb1dfce37b2d}\\
Workspace & 12 reference files: PDF drawings, XLSX and CSV data, Markdown constraints, and three OSHA booklets\\
Deliverable & \texttt{ME042\_RollFormer\_Guard\_Retrofit\_Review.xlsx}, five mandated sheets, formula-driven\\
Rubric & 42 items, 40 positive and 2 penalty; 98 positive points, 91 of them on required items\\
\end{tabularx}
\end{TaskMeta}

\begin{PromptBox}[Task A prompt $\cdot$ reproduced verbatim]
\textbf{Line 4 roll former guard retrofit decision review}\\[2pt]
Hi Priya,\\[3pt]

I need an engineer-ready decision file for the July 30 Line 4 design review. Last week's walkthrough found
an accessible guide-roll nip point and an exposed \#60 chain drive at the roll former infeed. Operations
still needs to thread 36--48 inch strip at the start of a coil, while maintenance needs a practical way to
service the rolls under lockout.

Please prepare one Excel workbook named 

\texttt{ME042\_RollFormer\_Guard\_Retrofit\_Review.xlsx}. This is
the file we will use to decide whether a guard concept can move into detailed design. Use the attached
current-state layout, measurements, observations, stop-test results, option estimates, and production
constraints. The OSHA booklets are background for the safeguarding and JHA reasoning; no fresh web research
is needed.

\textbf{Build exactly these five working tabs, with the decision summary first:}
\begin{enumerate}\setlength{\itemsep}{2pt}
\item \textbf{Decision Summary} --- state a clear recommendation for Concepts A, B, and C; identify the
selected concept, capital impact, lead time, highest remaining risk, and the conditions that must be met
before installation. Include a brief source note that distinguishes the site planning screen from a
regulatory determination.
\item \textbf{Risk \& JHA} --- turn the observation record into a usable JHA for threading, normal running,
debris clearing, and adjustment/service. Show initial and residual likelihood/severity/risk values,
engineered controls, the responsible verifier, and a release disposition for each item.
\item \textbf{Guard Geometry} --- document the relevant dimensions and calculations. Use the internal
interlocked-guard screen in the constraints file with the raw stop-test data, and show whether the current
hinged option has enough separation. Also check the fixed concept's feed opening, guard span, and
rotating-part clearance.
\item \textbf{Concept Comparison} --- compare all three concepts on personnel protection, production fit,
maintainability, budget, schedule, and verification burden. Use a weighted comparison and show the build-up
of each estimated cost, including contingency. A concise comparison chart is welcome.
\item \textbf{Action \& Verification} --- give the design-release, fabrication, fit-up, lockout, and startup
checks an owner, due date, evidence requirement, current status, and a clear no-go condition.
\end{enumerate}

Use formulas for calculated risk values, stop-distance screening, guard span, budget totals, capital
variance, and weighted scores rather than typing results. Keep the workbook readable in a design review:
freeze table headers, apply filters where helpful, and make holds or no-go items visually obvious. Cite the
relevant OSHA 3170 pages/sections in short notes where they support a control choice, but do not present
this workbook as a compliance determination. EHS and the machine-safety reviewer remain the final authority
for applicable requirements and validation.

Thanks,\\
Morgan Lee\\
Manufacturing Engineering Manager
\end{PromptBox}

\begin{EvidenceBox}[Task A attachments $\cdot$ 12 reference files]
The workspace mixes drawings, tabular data, planning notes, and regulatory background:
\texttt{ME042\_Current\_State\_Layout.pdf}, 

\texttt{ME042\_Design\_Basis.pdf},

\texttt{ME042\_Guard\_Measurements.xlsx}, 

\texttt{ME042\_Guard\_Concepts\_and\_Budget.xlsx},

\texttt{ME042\_JHA\_Worksheet\_Template.xlsx}, 

\texttt{ME042\_Operation\_Observation\_Log.csv},

\texttt{ME042\_Stop\_Test\_Data.csv}, 

\texttt{ME042\_Production\_and\_Maintenance\_Constraints.md},

\texttt{ME042\_Review\_Roles\_and\_Release\_Gates.md}, 

\texttt{OSHA\_3071\_Job\_Hazard\_Analysis.pdf},

\texttt{OSHA\_3080\_Hand\_and\_Power\_Tools.pdf}

\texttt{OSHA\_3170\_Safeguarding\_Equipment\_and\_Protecting\_Employees\_from\_Amputations.pdf}.

\begin{center}
\includegraphics[width=\linewidth]{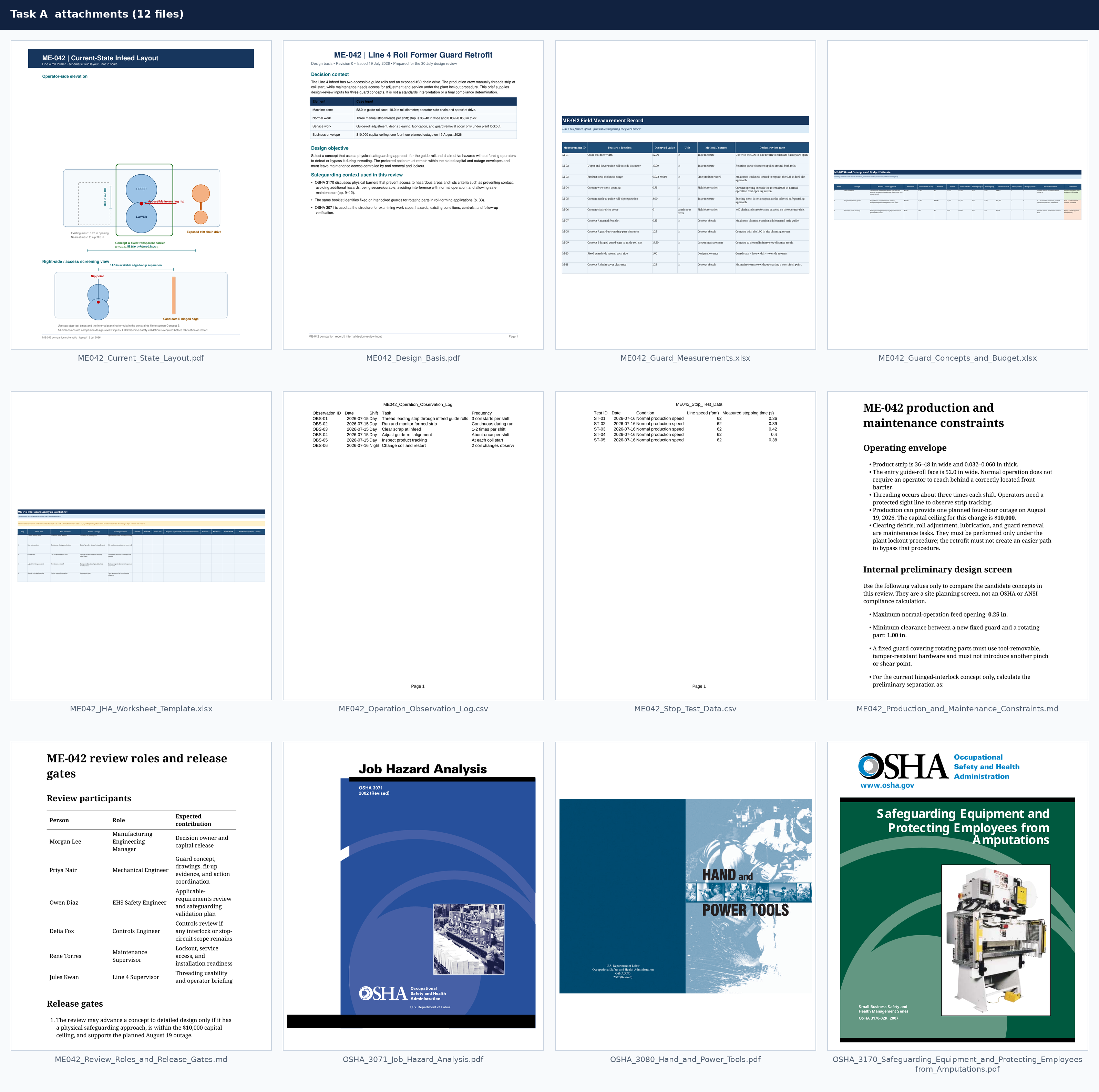}
\end{center}
\evidencecaption{Montage of all 12 attachments. Spreadsheets, logs, and Markdown notes appear as their rendered first pages; PDFs as their first page.}

\begin{center}
\includegraphics[width=\linewidth]{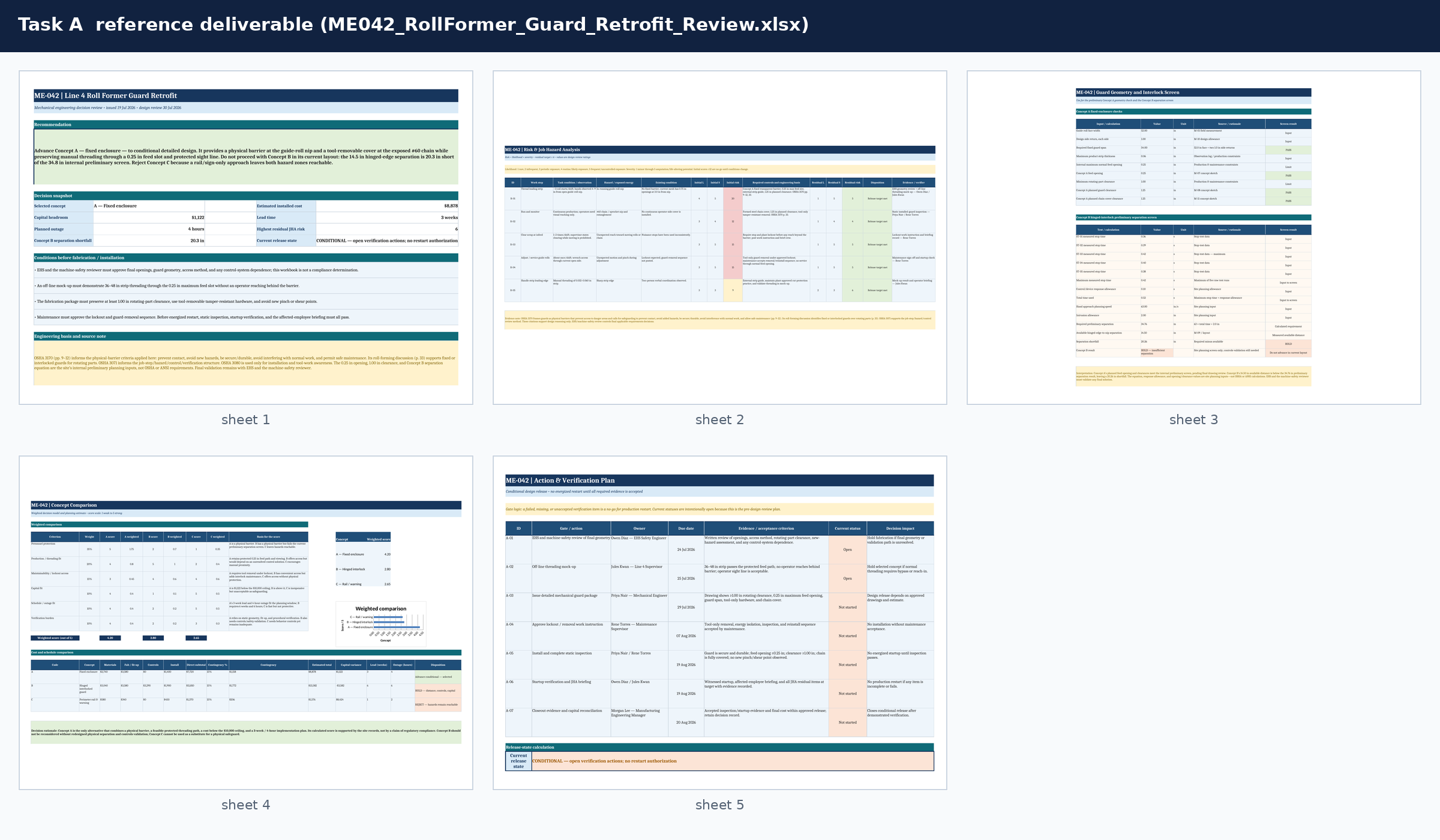}
\end{center}
\evidencecaption{Reference workbook montage: the five requested working sheets in decision-summary order.}
\end{EvidenceBox}

\begin{RubricBox}[Task A rubric $\cdot$ acceptance contract]
Positive points total \textbf{98}. The \textbf{33} required items carry \textbf{91} of those points, and the remaining \textbf{7} points sit on optional items. The \textbf{2} negative rows are penalties, applied only when their condition occurs; the required points are also the pass threshold, so the rubric cannot be cleared as a flat checklist.
\smallskip
\noindent\colorbox{themebg}{\parbox{\dimexpr\linewidth-2\fboxsep\relax}{\textbf{\small Deliverable and workbook structure}}}
\par\nobreak\smallskip
\begin{tabularx}{\linewidth}{@{}>{\raggedright\arraybackslash}p{.06\linewidth}@{\hspace{6pt}}X>{\raggedleft\arraybackslash}p{.07\linewidth}>{\centering\arraybackslash}p{.135\linewidth}@{}}
\toprule
\textbf{ID} & \textbf{Criterion} & \textbf{Pts} & \textbf{Status}\\
\midrule
\texttt{R01} & Delivers a readable Excel workbook named ME042\_RollFormer\_Guard\_Retrofit\_Review.xlsx. & +2 & \textcolor{accent}{Required}\\
\texttt{R02} & The workbook has exactly the five requested working sheets: Decision Summary, Risk \& JHA, Guard Geometry, Concept Comparison, and Action \& Verification. & +4 & \textcolor{accent}{Required}\\
\texttt{R03} & Does not add a separate source, scratch, or hidden working sheet; source notes are integrated into the requested review tabs. & +1 & \textcolor{muted}{Optional}\\
\bottomrule
\end{tabularx}
\medskip
\noindent\colorbox{themebg}{\parbox{\dimexpr\linewidth-2\fboxsep\relax}{\textbf{\small Decision summary and source framing}}}
\par\nobreak\smallskip
\begin{tabularx}{\linewidth}{@{}>{\raggedright\arraybackslash}p{.06\linewidth}@{\hspace{6pt}}X>{\raggedleft\arraybackslash}p{.07\linewidth}>{\centering\arraybackslash}p{.135\linewidth}@{}}
\toprule
\textbf{ID} & \textbf{Criterion} & \textbf{Pts} & \textbf{Status}\\
\midrule
\texttt{R04} & Decision Summary makes an unambiguous recommendation to advance Concept A only conditionally into detailed design. & +4 & \textcolor{accent}{Required}\\
\texttt{R05} & The recommended Concept A is described as a fixed physical barrier with a protected feed path, separate chain coverage, and tool-only removal for maintenance rather than as a generic guard. & +4 & \textcolor{accent}{Required}\\
\texttt{R06} & Explains that Concept B cannot advance in its current layout because the 14.5 in available separation is about 20.3 in short of the preliminary screen (required separation about 34.8 in). & +4 & \textcolor{accent}{Required}\\
\texttt{R07} & Rejects Concept C specifically because a rail/sign/procedure-only approach leaves the guide-roll and chain hazards physically reachable. & +2 & \textcolor{accent}{Required}\\
\texttt{R08} & Decision snapshot shows the selected option's estimated installed cost, capital headroom, lead time, planned outage, and highest residual risk in visible decision-ready form. & +3 & \textcolor{accent}{Required}\\
\texttt{R09} & Identifies the expected Concept A capital headroom as about \$1,122 against the \$10,000 ceiling. & +1 & \textcolor{muted}{Optional}\\
\texttt{R10} & Includes a concise source/assumption note that cites OSHA 3170's machine-safeguarding discussion and clearly separates the site planning screen from a compliance determination. & +3 & \textcolor{accent}{Required}\\
\bottomrule
\end{tabularx}
\medskip
\noindent\colorbox{themebg}{\parbox{\dimexpr\linewidth-2\fboxsep\relax}{\textbf{\small Risk and job hazard analysis}}}
\par\nobreak\smallskip
\begin{tabularx}{\linewidth}{@{}>{\raggedright\arraybackslash}p{.06\linewidth}@{\hspace{6pt}}X>{\raggedleft\arraybackslash}p{.07\linewidth}>{\centering\arraybackslash}p{.135\linewidth}@{}}
\toprule
\textbf{ID} & \textbf{Criterion} & \textbf{Pts} & \textbf{Status}\\
\midrule
\texttt{R11} & Risk \& JHA covers all five relevant work activities: threading, running/monitoring, debris clearing, adjustment/service, and sharp-edge handling. & +3 & \textcolor{accent}{Required}\\
\texttt{R12} & The threading row recognizes the accessible in-running guide-roll nip and starts at likelihood 4, severity 5, initial risk 20 (or an equally explicit 4$\times$5 calculation). & +2 & \textcolor{accent}{Required}\\
\texttt{R13} & The continuous-running row treats the exposed \#60 chain/sprocket drive as a separate entanglement/nip hazard and assigns a physical cover control. & +2 & \textcolor{accent}{Required}\\
\texttt{R14} & Debris clearing and adjustment/service are explicitly tied to stopping and the plant lockout process; the workbook does not normalize reach-in work while moving. & +2 & \textcolor{accent}{Required}\\
\texttt{R15} & Addresses the sharp leading-edge hazard during manual strip handling with an external guide and the plant-approved cut-protection practice or an equally specific control. & +1 & \textcolor{muted}{Optional}\\
\texttt{R16} & Initial and residual risk values are formula-driven from likelihood and severity, not merely typed as unexplained totals. & +3 & \textcolor{accent}{Required}\\
\texttt{R17} & Each residual JHA item has a visible disposition, and the proposed residual risks meet the internal target of 6 or below without hiding the original high-risk conditions. & +2 & \textcolor{accent}{Required}\\
\texttt{R18} & JHA evidence/verification fields name an accountable reviewer or evidence source for the guard geometry, chain cover, lockout, maintenance, and threading controls. & +3 & \textcolor{accent}{Required}\\
\bottomrule
\end{tabularx}
\medskip
\noindent\colorbox{themebg}{\parbox{\dimexpr\linewidth-2\fboxsep\relax}{\textbf{\small Guard geometry and preliminary screening}}}
\par\nobreak\smallskip
\begin{tabularx}{\linewidth}{@{}>{\raggedright\arraybackslash}p{.06\linewidth}@{\hspace{6pt}}X>{\raggedleft\arraybackslash}p{.07\linewidth}>{\centering\arraybackslash}p{.135\linewidth}@{}}
\toprule
\textbf{ID} & \textbf{Criterion} & \textbf{Pts} & \textbf{Status}\\
\midrule
\texttt{R19} & Guard Geometry calculates the fixed guard span as 54.0 in from the 52.0 in face width plus two 1.0 in side returns. & +2 & \textcolor{accent}{Required}\\
\texttt{R20} & Checks the Concept A 0.25 in feed opening against the internal 0.25 in maximum and marks the result clearly. & +2 & \textcolor{accent}{Required}\\
\texttt{R21} & Checks both the planned front-barrier clearance and chain-cover clearance of 1.25 in against the 1.00 in minimum planning screen. & +2 & \textcolor{accent}{Required}\\
\texttt{R22} & Lists the five raw stop-test values and calculates the maximum measured stop time as 0.42 s using a formula. & +2 & \textcolor{accent}{Required}\\
\texttt{R23} & Adds the 0.10 s site response allowance and calculates the total time used as 0.52 s. & +1 & \textcolor{muted}{Optional}\\
\texttt{R24} & Applies the supplied preliminary equation 63 $\times$ (maximum stop time + 0.10 s) + 2.0 and calculates a required separation of about 34.76 in. & +4 & \textcolor{accent}{Required}\\
\texttt{R25} & Compares the 34.76 in requirement with the 14.50 in available distance, calculates the roughly 20.26 in shortfall, and visibly holds Concept B. & +3 & \textcolor{accent}{Required}\\
\texttt{R26} & Labels the separation equation, opening limit, and clearance limit as internal preliminary planning inputs rather than OSHA or ANSI requirements. & +2 & \textcolor{accent}{Required}\\
\bottomrule
\end{tabularx}
\medskip
\noindent\colorbox{themebg}{\parbox{\dimexpr\linewidth-2\fboxsep\relax}{\textbf{\small Concept comparison and budget}}}
\par\nobreak\smallskip
\begin{tabularx}{\linewidth}{@{}>{\raggedright\arraybackslash}p{.06\linewidth}@{\hspace{6pt}}X>{\raggedleft\arraybackslash}p{.07\linewidth}>{\centering\arraybackslash}p{.135\linewidth}@{}}
\toprule
\textbf{ID} & \textbf{Criterion} & \textbf{Pts} & \textbf{Status}\\
\midrule
\texttt{R27} & Concept Comparison scores all three concepts on personnel protection, production/threading fit, maintainability/lockout access, capital fit, schedule/outage fit, and verification burden. & +3 & \textcolor{accent}{Required}\\
\texttt{R28} & Uses the stated weighted model (35\%, 20\%, 15\%, 10\%, 10\%, 10\%) with formulas for the weighted result rather than a purely narrative ranking. & +3 & \textcolor{accent}{Required}\\
\texttt{R29} & The weighted outcome visibly ranks Concept A above both B and C and gives a technical basis for the lower B and C ratings. & +2 & \textcolor{accent}{Required}\\
\texttt{R30} & Shows cost build-up by materials, fabrication/fit-up, controls, installation, direct subtotal, 15\% contingency, estimated total, capital variance, lead time, and outage for every concept. & +3 & \textcolor{accent}{Required}\\
\texttt{R31} & Cost totals are correct or materially equivalent: Concept A about \$8,878, Concept B about \$13,582, and Concept C about \$1,576. & +3 & \textcolor{accent}{Required}\\
\texttt{R32} & Makes clear that Concept A fits the \$10,000 ceiling and four-hour outage, while B exceeds both the capital and planned-outage constraints. & +2 & \textcolor{accent}{Required}\\
\texttt{R33} & States a distinct disposition for each concept: advance Concept A conditionally, hold B pending redesigned separation/controls validation, and reject C for inadequate physical safeguarding. & +2 & \textcolor{accent}{Required}\\
\bottomrule
\end{tabularx}
\medskip
\noindent\colorbox{themebg}{\parbox{\dimexpr\linewidth-2\fboxsep\relax}{\textbf{\small Action plan, release gate, and usability}}}
\par\nobreak\smallskip
\begin{tabularx}{\linewidth}{@{}>{\raggedright\arraybackslash}p{.06\linewidth}@{\hspace{6pt}}X>{\raggedleft\arraybackslash}p{.07\linewidth}>{\centering\arraybackslash}p{.135\linewidth}@{}}
\toprule
\textbf{ID} & \textbf{Criterion} & \textbf{Pts} & \textbf{Status}\\
\midrule
\texttt{R34} & Action \& Verification contains named owners, due dates, and acceptance evidence for EHS geometry review, threading mock-up, detailed drawings, lockout/removal instruction, static installation inspection, startup/JHA briefing, and closeout. & +4 & \textcolor{accent}{Required}\\
\texttt{R35} & The action plan uses the relevant people/roles from the review record, including EHS (Owen Diaz), mechanical engineering (Priya Nair), maintenance (Rene Torres), operations (Jules Kwan), and management (Morgan Lee), or clearly equivalent role assignments. & +3 & \textcolor{accent}{Required}\\
\texttt{R36} & Static inspection acceptance includes the key measurable checks: feed opening at or below 0.25 in, clearance at or above 1.00 in, fully covered chain drive, secure tool-only hardware, and no new pinch/shear point. & +3 & \textcolor{accent}{Required}\\
\texttt{R37} & A visible release-state/no-go rule prevents energized production restart until required verification is complete and accepted. & +3 & \textcolor{accent}{Required}\\
\texttt{R38} & Tables are usable in a live design review: headers are frozen, key records are filterable, calculations have sensible formats, and holds/no-go items are visually distinguishable. & +1 & \textcolor{muted}{Optional}\\
\texttt{R39} & Concept Comparison includes a clear visual comparison (such as a weighted-score chart) that supports rather than replaces the numerical table. & +1 & \textcolor{muted}{Optional}\\
\texttt{R40} & The workbook is consistently formatted, readable at normal zoom, and has no broken references, formula errors, clipped central content, or unexplained blank decision fields. & +1 & \textcolor{muted}{Optional}\\
\bottomrule
\end{tabularx}
\medskip
\noindent\colorbox{themebg}{\parbox{\dimexpr\linewidth-2\fboxsep\relax}{\textbf{\small Penalty conditions}}}
\par\nobreak\smallskip
\begin{tabularx}{\linewidth}{@{}>{\raggedright\arraybackslash}p{.06\linewidth}@{\hspace{6pt}}X>{\raggedleft\arraybackslash}p{.07\linewidth}>{\centering\arraybackslash}p{.135\linewidth}@{}}
\toprule
\textbf{ID} & \textbf{Criterion} & \textbf{Pts} & \textbf{Status}\\
\midrule
\rowcolor{penaltytint}
\texttt{R41} & Subtract if the workbook claims OSHA compliance, OSHA certification, or final regulatory approval instead of retaining EHS/machine-safety validation as a condition. & \textcolor{penalty}{\textbf{-5}} & \textcolor{penalty}{Penalty}\\
\rowcolor{penaltytint}
\texttt{R42} & Subtract if it relies on fabricated external citations or replaces the supplied site planning inputs with unsupported safety-distance, opening, or clearance rules. & \textcolor{penalty}{\textbf{-3}} & \textcolor{penalty}{Penalty}\\
\bottomrule
\end{tabularx}
\medskip
\end{RubricBox}
\clearpage

\subsection{Task B $\cdot$ Art Opens a Place oral-history highlight}
\label{app:task-b}

\begin{TaskMeta}[Task B $\cdot$ single-source video-editing task]
\begin{tabularx}{\linewidth}{@{}>{\bfseries}p{.205\linewidth}X@{}}
Occupation & Film and Video Editors\\
Work request & NPS Acadia oral-history web highlight for the refreshed Shirley Beccue feature page\\
Task identifier & \texttt{11390cf2-1d7a-46cf-99a6-a7029ea05332}\\
Workspace & 1 source video \texttt{nps\_acadia\_shirley\_beccue\_source.mp4} (854$\times$480, 23.976 fps, H.264/AAC, 220 s)\\
Deliverable & \texttt{art\_opens\_a\_place.mp4}, a 24--28 s 16:9 cut with burned-in captions\\
Rubric & 43 items, 37 positive and 6 penalty; 121 positive points, 117 of them on required items\\
\end{tabularx}
\end{TaskMeta}

\begin{PromptBox}[Task B prompt $\cdot$ reproduced verbatim]
Hi,

We are refreshing the legacy feature page this week and need a compact video that helps web visitors decide
to watch the complete Shirley Beccue interview. The attached
\texttt{nps\_acadia\_shirley\_beccue\_source.mp4} is the only approved picture and sound. The audience is a
general visitor arriving cold on the Acadia feature page, so the tone should be reflective, human, and easy
to follow without sound.

Please deliver one finished file named \texttt{art\_opens\_a\_place.mp4}. Make a 24--28 second 16:9 highlight
that turns Beccue's reflection on art into one complete thought. Build the spoken spine around her point that
art \textquotedblleft{}really touches your emotions\textquotedblright{} and retain her complete line,
\textquotedblleft{}It invites you into a place that you might not otherwise go, whether it's painting, music
or dance.\textquotedblright{} It should resolve cleanly rather than trail off. You may remove pauses and use
the Artist-in-Residence cutaways already in the interview, but do not rearrange her words, create new speech,
or imply that a person shown in a cutaway is speaking.

Please leave out the early material about women in Park Service uniforms and the later web-training story;
both are good material, but neither belongs in this feature. Use only the supplied video for picture and
sound---no added music, narration, stock footage, logos, or new factual claims. Source-contained labels may
remain when a cutaway genuinely earns its place.

Open with the exact title \texttt{ART OPENS A PLACE} within the first two seconds. When the main interview
shot first appears, use this identification exactly:

\begin{quote}\small
Shirley Beccue\\
Former Assistant Chief of Interpretation, Acadia National Park
\end{quote}

Burn in accurate, sentence-case captions for all spoken words. Keep them to two lines or fewer, readable
against the image, and clear of Beccue's face and the identification. The finish should hold this exact
closing copy long enough to read; it may wrap across lines:

\begin{quote}\small
Legacy: One Woman's Continuing Influence on Acadia National Park\\
Watch the full interview
\end{quote}

Keep the look restrained and native to the interview. Use clean, motivated edits, preserve the source's
proportions, and make sure the dialogue is intelligible and in sync. Export progressive H.264 video at
854$\times$480, 23.976 fps (24000/1001), yuv420p, with 48 kHz stereo AAC audio in an MP4 container. Do not
include a project file, alternate version, or separate caption file.
\end{PromptBox}

\begin{EvidenceBox}[Task B attachment and visual evidence]
\begin{center}
\includegraphics[width=\linewidth]{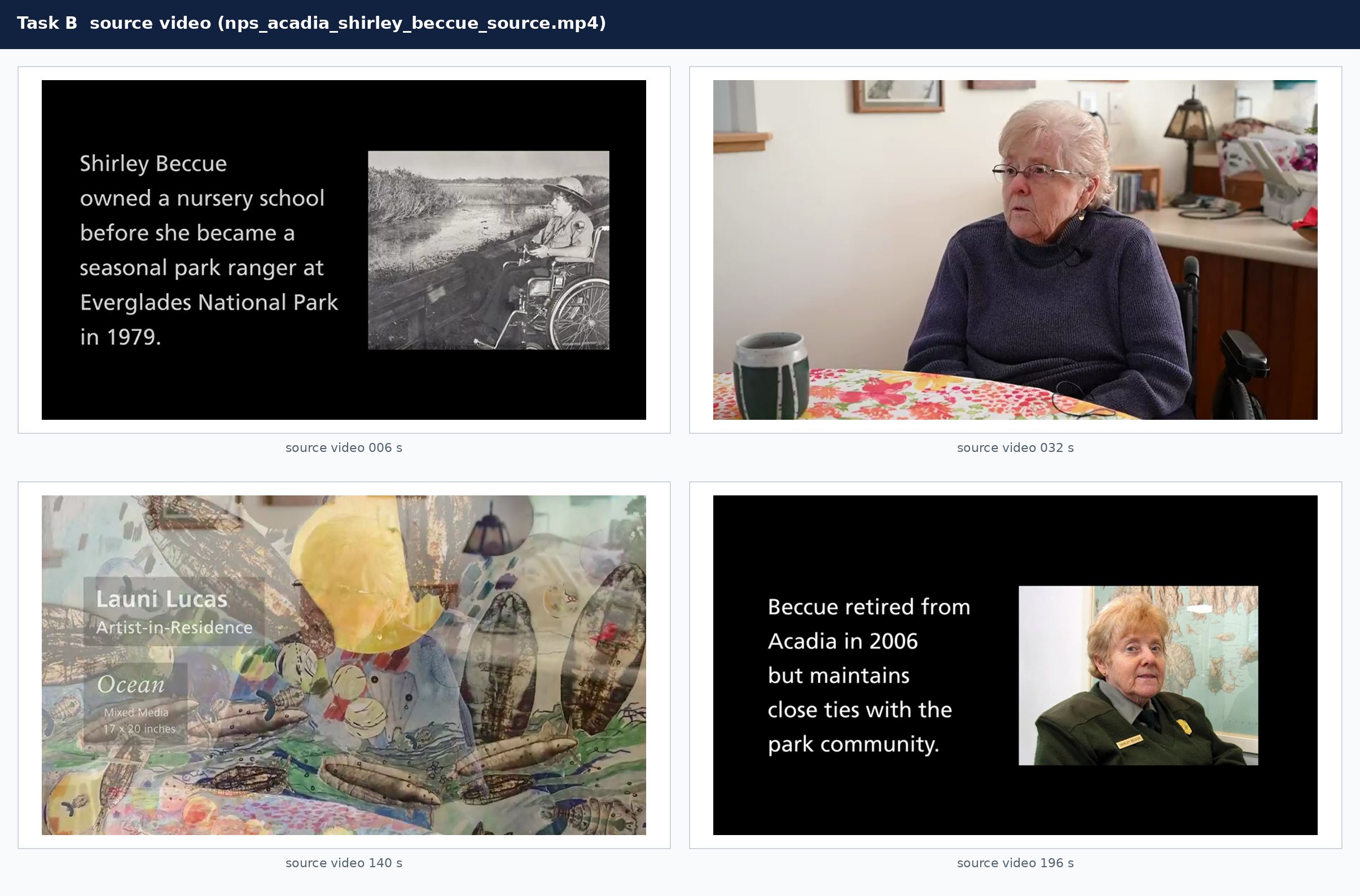}
\end{center}
\evidencecaption{Source montage sampled at 6 s, 32 s, 140 s, and 196 s. The third frame is the Artist-in-Residence cutaway the brief permits.}

\begin{center}
\includegraphics[width=\linewidth]{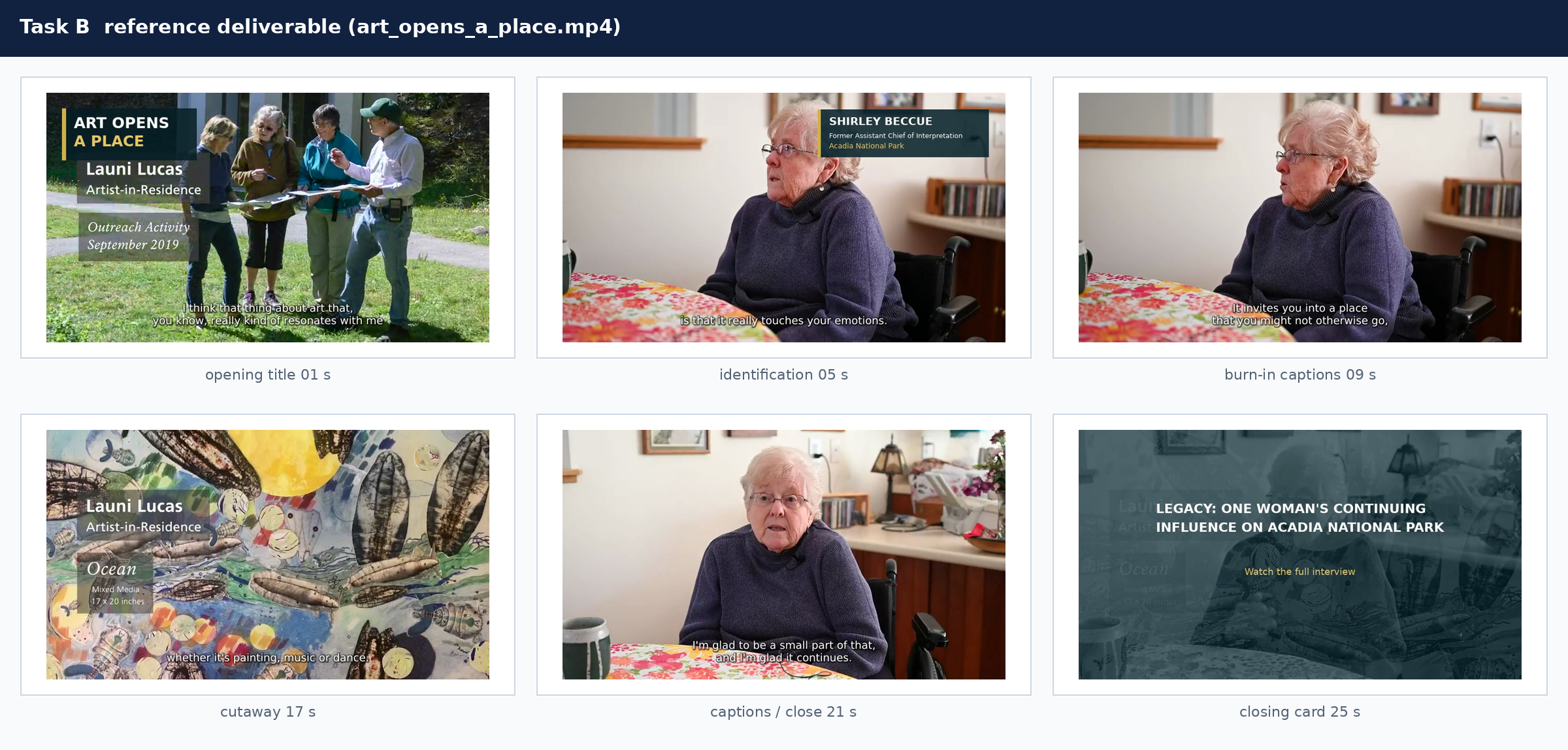}
\end{center}
\evidencecaption{Reference export sampled at 1 s, 5 s, 9 s, 17 s, 21 s, and 25 s: opening title, role identification, burned-in captions, the Artist-in-Residence cutaway, closing captions, and the closing card.}
\end{EvidenceBox}

\begin{RubricBox}[Task B rubric $\cdot$ acceptance contract]
Positive points total \textbf{121}. The \textbf{35} required items carry \textbf{117} of those points, and the remaining \textbf{4} points sit on optional items. The \textbf{6} negative rows are penalties, applied only when their condition occurs; the required points are also the pass threshold, so the rubric cannot be cleared as a flat checklist.
\smallskip
\noindent\colorbox{themebg}{\parbox{\dimexpr\linewidth-2\fboxsep\relax}{\textbf{\small Deliverable and technical delivery}}}
\par\nobreak\smallskip
\begin{tabularx}{\linewidth}{@{}>{\raggedright\arraybackslash}p{.06\linewidth}@{\hspace{6pt}}X>{\raggedleft\arraybackslash}p{.07\linewidth}>{\centering\arraybackslash}p{.135\linewidth}@{}}
\toprule
\textbf{ID} & \textbf{Criterion} & \textbf{Pts} & \textbf{Status}\\
\midrule
\texttt{R01} & The submission contains one playable final video named art\_opens\_a\_place.mp4. & +5 & \textcolor{accent}{Required}\\
\texttt{R03} & The final file is an MP4 container that opens and plays through without a decoding error. & +2 & \textcolor{accent}{Required}\\
\texttt{R04} & Measured runtime, including the opening and closing graphics, is at least 24.0 seconds and no more than 28.0 seconds. & +4 & \textcolor{accent}{Required}\\
\texttt{R05} & The delivered raster is exactly 854$\times$480 pixels. & +3 & \textcolor{accent}{Required}\\
\texttt{R06} & The finished frame is 16:9 and does not stretch the supplied footage. & +2 & \textcolor{accent}{Required}\\
\texttt{R07} & Video timing is 24000/1001 fps (23.976 fps), matching the supplied master. & +2 & \textcolor{accent}{Required}\\
\texttt{R08} & The video stream is progressive H.264 with yuv420p pixel format. & +3 & \textcolor{accent}{Required}\\
\texttt{R09} & The audio stream is 48 kHz stereo AAC. & +2 & \textcolor{accent}{Required}\\
\texttt{R10} & The first visible frame presents active source imagery or intentional opening graphics; it is not an accidental black, slate, or frozen frame. & +2 & \textcolor{accent}{Required}\\
\bottomrule
\end{tabularx}
\medskip
\noindent\colorbox{themebg}{\parbox{\dimexpr\linewidth-2\fboxsep\relax}{\textbf{\small Locked graphics and closing}}}
\par\nobreak\smallskip
\begin{tabularx}{\linewidth}{@{}>{\raggedright\arraybackslash}p{.06\linewidth}@{\hspace{6pt}}X>{\raggedleft\arraybackslash}p{.07\linewidth}>{\centering\arraybackslash}p{.135\linewidth}@{}}
\toprule
\textbf{ID} & \textbf{Criterion} & \textbf{Pts} & \textbf{Status}\\
\midrule
\texttt{R11} & Within the first two seconds, the exact opening title \textquotedblleft{}ART OPENS A PLACE\textquotedblright{} is fully legible on screen. & +6 & \textcolor{accent}{Required}\\
\texttt{R12} & The opening title has adequate contrast, sensible line breaks, and enough dwell time to be read without obscuring the key visual action. & +2 & \textcolor{accent}{Required}\\
\texttt{R13} & On Shirley Beccue's first on-camera interview shot, the identification reads exactly \textquotedblleft{}Shirley Beccue\textquotedblright{} and \textquotedblleft{}Former Assistant Chief of Interpretation, Acadia National Park.\textquotedblright{} & +5 & \textcolor{accent}{Required}\\
\texttt{R14} & The identification is clearly hierarchical and does not collide with Beccue's face or the burned-in captions. & +2 & \textcolor{accent}{Required}\\
\texttt{R15} & The closing card carries the exact title \textquotedblleft{}Legacy: One Woman's Continuing Influence on Acadia National Park\textquotedblright{} and the exact invitation \textquotedblleft{}Watch the full interview\textquotedblright{}; line wrapping may vary. & +5 & \textcolor{accent}{Required}\\
\texttt{R16} & The closing copy remains legible for at least two seconds and the video ends immediately after an intentional close or fade rather than an accidental hold. & +2 & \textcolor{accent}{Required}\\
\bottomrule
\end{tabularx}
\medskip
\noindent\colorbox{themebg}{\parbox{\dimexpr\linewidth-2\fboxsep\relax}{\textbf{\small Editorial story and source fidelity}}}
\par\nobreak\smallskip
\begin{tabularx}{\linewidth}{@{}>{\raggedright\arraybackslash}p{.06\linewidth}@{\hspace{6pt}}X>{\raggedleft\arraybackslash}p{.07\linewidth}>{\centering\arraybackslash}p{.135\linewidth}@{}}
\toprule
\textbf{ID} & \textbf{Criterion} & \textbf{Pts} & \textbf{Status}\\
\midrule
\texttt{R17} & The spoken editorial spine is a coherent reflection on art that includes Beccue's point that art \textquotedblleft{}really touches your emotions\textquotedblright{} and her complete line \textquotedblleft{}It invites you into a place that you might not otherwise go, whether it's painting, music or dance.\textquotedblright{} & +9 & \textcolor{accent}{Required}\\
\texttt{R18} & The word sequence through \textquotedblleft{}whether it's painting, music or dance\textquotedblright{} is presented as one intelligible thought, with no missing clause that changes its meaning. & +3 & \textcolor{accent}{Required}\\
\texttt{R19} & The edit resolves the reflection into a complete ending rather than cutting away in the middle of a sentence or leaving the viewer without a final thought. & +4 & \textcolor{accent}{Required}\\
\texttt{R20} & No selected material comes from the source's early discussion of women in Park Service uniforms or its web-training/web-coding story. & +4 & \textcolor{accent}{Required}\\
\texttt{R21} & Dialogue is not reordered, synthetically altered, or edited to make Beccue appear to state a claim she did not make. & +5 & \textcolor{accent}{Required}\\
\bottomrule
\end{tabularx}
\medskip
\noindent\colorbox{themebg}{\parbox{\dimexpr\linewidth-2\fboxsep\relax}{\textbf{\small Captions, sound, and synchronization}}}
\par\nobreak\smallskip
\begin{tabularx}{\linewidth}{@{}>{\raggedright\arraybackslash}p{.06\linewidth}@{\hspace{6pt}}X>{\raggedleft\arraybackslash}p{.07\linewidth}>{\centering\arraybackslash}p{.135\linewidth}@{}}
\toprule
\textbf{ID} & \textbf{Criterion} & \textbf{Pts} & \textbf{Status}\\
\midrule
\texttt{R22} & Every audible spoken phrase in the final cut has burned-in captions. & +4 & \textcolor{accent}{Required}\\
\texttt{R23} & Caption wording faithfully reflects the selected speech, including \textquotedblleft{}touches your emotions\textquotedblright{} and \textquotedblleft{}whether it's painting, music or dance,\textquotedblright{} without invented completion or altered meaning. & +3 & \textcolor{accent}{Required}\\
\texttt{R24} & Captions use sentence case, readable punctuation, and no more than two lines at a time. & +2 & \textcolor{accent}{Required}\\
\texttt{R25} & Caption contrast, outline or backing, and safe placement make each line readable while keeping it clear of the speaker and the role identification. & +3 & \textcolor{accent}{Required}\\
\texttt{R26} & Source dialogue is consistently intelligible at a natural listening level. & +4 & \textcolor{accent}{Required}\\
\texttt{R29} & Where Beccue is shown speaking, lip sync is natural; any voice-over on a source cutaway is clearly intentional and does not falsely attribute speech to another person. & +4 & \textcolor{accent}{Required}\\
\texttt{R30} & Dialogue edits preserve full words, avoid audible clicks or pops, and do not create an unexplained silent gap inside the spoken thought. & +3 & \textcolor{accent}{Required}\\
\bottomrule
\end{tabularx}
\medskip
\noindent\colorbox{themebg}{\parbox{\dimexpr\linewidth-2\fboxsep\relax}{\textbf{\small Cutaways and picture finishing}}}
\par\nobreak\smallskip
\begin{tabularx}{\linewidth}{@{}>{\raggedright\arraybackslash}p{.06\linewidth}@{\hspace{6pt}}X>{\raggedleft\arraybackslash}p{.07\linewidth}>{\centering\arraybackslash}p{.135\linewidth}@{}}
\toprule
\textbf{ID} & \textbf{Criterion} & \textbf{Pts} & \textbf{Status}\\
\midrule
\texttt{R31} & If an Artist-in-Residence cutaway is used, it advances the art theme and is visually distinct from the interview rather than serving as filler. & +2 & \textcolor{muted}{Optional}\\
\texttt{R32} & Any source-contained Artist-in-Residence labels are left as part of the original footage and are not reframed into a new factual claim about Beccue or the closing feature. & +3 & \textcolor{accent}{Required}\\
\texttt{R33} & Reframing preserves the supplied footage's proportions and keeps important facial and visual detail in frame without obvious stretching, pillarboxing, or accidental crop damage. & +3 & \textcolor{accent}{Required}\\
\texttt{R34} & Text and graphics stay inside safe margins with no clipped letters, edge crowding, or unintended overlap. & +2 & \textcolor{accent}{Required}\\
\texttt{R35} & Color, exposure, sharpness, and scale remain visually coherent across interview and cutaway shots. & +2 & \textcolor{muted}{Optional}\\
\texttt{R36} & Transitions are restrained and motivated; the video contains no distracting wipes, glitches, or effect-driven cuts. & +2 & \textcolor{accent}{Required}\\
\bottomrule
\end{tabularx}
\medskip
\noindent\colorbox{themebg}{\parbox{\dimexpr\linewidth-2\fboxsep\relax}{\textbf{\small Structure, tone, and audience fit}}}
\par\nobreak\smallskip
\begin{tabularx}{\linewidth}{@{}>{\raggedright\arraybackslash}p{.06\linewidth}@{\hspace{6pt}}X>{\raggedleft\arraybackslash}p{.07\linewidth}>{\centering\arraybackslash}p{.135\linewidth}@{}}
\toprule
\textbf{ID} & \textbf{Criterion} & \textbf{Pts} & \textbf{Status}\\
\midrule
\texttt{R37} & The cut moves from a quick thematic opening to the reflection and then the invitation at a pace appropriate for a visitor encountering the page for the first time. & +3 & \textcolor{accent}{Required}\\
\texttt{R38} & The overall tone is warm, reflective, and human rather than promotional, sensational, or overly stylized. & +3 & \textcolor{accent}{Required}\\
\texttt{R39} & Graphics use a restrained hierarchy that is legible at web-player size and supports, rather than competes with, the source imagery. & +3 & \textcolor{accent}{Required}\\
\texttt{R40} & No added dates, program names, biographical facts, attributions, or other factual claims appear beyond the locked identification, closing copy, accurate captions, and source-contained labels. & +3 & \textcolor{accent}{Required}\\
\bottomrule
\end{tabularx}
\medskip
\noindent\colorbox{themebg}{\parbox{\dimexpr\linewidth-2\fboxsep\relax}{\textbf{\small Penalty conditions}}}
\par\nobreak\smallskip
\begin{tabularx}{\linewidth}{@{}>{\raggedright\arraybackslash}p{.06\linewidth}@{\hspace{6pt}}X>{\raggedleft\arraybackslash}p{.07\linewidth}>{\centering\arraybackslash}p{.135\linewidth}@{}}
\toprule
\textbf{ID} & \textbf{Criterion} & \textbf{Pts} & \textbf{Status}\\
\midrule
\rowcolor{penaltytint}
\texttt{R02} & Deduct for any submitted project file, alternate cut, separate caption file, or other unrequested deliverable. & \textcolor{penalty}{\textbf{-3}} & \textcolor{penalty}{Penalty}\\
\rowcolor{penaltytint}
\texttt{R27} & Deduct if added music, narration, sound effects, or any other non-source audio appears in the cut. & \textcolor{penalty}{\textbf{-8}} & \textcolor{penalty}{Penalty}\\
\rowcolor{penaltytint}
\texttt{R28} & Deduct if stock footage, externally sourced imagery, generative imagery, or a newly created logo is used instead of the supplied video. & \textcolor{penalty}{\textbf{-8}} & \textcolor{penalty}{Penalty}\\
\rowcolor{penaltytint}
\texttt{R41} & Deduct for any misspelling or wording change in the required opening title, role identification, or closing copy. & \textcolor{penalty}{\textbf{-5}} & \textcolor{penalty}{Penalty}\\
\rowcolor{penaltytint}
\texttt{R42} & Deduct for a visible watermark, NPS Arrowhead symbol, technical overlay, source filename, placeholder, or unrelated logo. & \textcolor{penalty}{\textbf{-5}} & \textcolor{penalty}{Penalty}\\
\rowcolor{penaltytint}
\texttt{R43} & Deduct if the export is corrupted, missing either a usable video or audio stream, or materially fails the requested technical delivery specification. & \textcolor{penalty}{\textbf{-6}} & \textcolor{penalty}{Penalty}\\
\bottomrule
\end{tabularx}
\medskip
\end{RubricBox}
\clearpage


\section{Prompts}
\label{app:prompts}
In this section, we present in detail the prompts used by each module of WorkGenesis.

\subsection{Stage 1: Discovering Real-World Anchors}

\begin{wgpromptbox}[Stage 1 -- Discover real-world anchors]
You are an occupational research specialist. You build the factual foundation for a new professional work task by finding real documents on the public web, downloading them, inspecting them with code, and deciding whether they are strong enough to anchor a realistic task. You alternate searching and judging until you have a usable set of real files, then you record what you found. End your run with the JSON summary described below.

# Task

{INSTRUCTION}

# Occupation playbook

{OCCUPATION_PLAYBOOK}

The playbook describes the occupation, its sector, its duties, the work activities it covers, and the tools and documents it uses. Use it to decide which files a real practitioner in this occupation would handle and which documents would realistically appear at the desk. Its example directions are a menu, so choose the document that the public material you can actually retrieve best supports.

# Lineage memory

{LINEAGE_MEMORY}

This lists anchors that earlier tasks in the same lineage already used, with their source page URLs, direct file URLs, domains, and SHA-256 hashes, along with short summaries of the scenarios those tasks built. Before you search, read it. Prefer a different source family from the ones already used when a good alternative exists. A domain a sibling task used may be revisited only for a different file, and a direct file URL a sibling task used is never reused. A file you keep must not match anything here by direct URL or by hash.

# Required file formats

{REQUIRED_FILE_FORMATS}

# Workspace

Work inside {WORKSPACE_DIR}. Keep every real file you decide to use in {WORKSPACE_DIR}/anchors/.

# Tools

- web_search: runs a search query and returns ranked results with title, URL, and snippet.
- web_fetch: opens a URL, extracts its readable content, and downloads binary files to the workspace.
- code_exec: runs shell or Python commands in a sandbox for inspecting downloaded files.
- view_image: shows an image file from the workspace, including scanned pages, screenshots, and chart renders.

# Instructions

## What you produce

A small set of real files in {WORKSPACE_DIR}/anchors/, each suitable to anchor a professional task, together with {WORKSPACE_DIR}/anchors/metadata.json recording where each file came from and why it was accepted. At least one usable real file is required.

## Suitability checks

Judge every candidate against all of the checks below and record each judgment. Accept a file only when every check passes.

- Occupation fit: a worker in the stated occupation would use this file.
- Format fit: the file type matches {REQUIRED_FILE_FORMATS} when that is not empty. Working formats are pdf, docx, xlsx, csv, pptx, and their close variants.
- Substance: the file is non-trivial and populated, with enough records, tables, clauses, figures, or fields to support a real analysis or document task. Reject blank templates, one-line stubs, cover pages, login walls, and marketing splash pages.
- Veracity: the file is a real artifact from a credible source, and you have opened and read it. Reject AI-generated filler, placeholders, and files that failed to download or decompress.
- Novelty: the direct file URL and the SHA-256 both stay clear of the lineage memory, and the source family is new where a good alternative exists. Compute the hash and compare before you accept the file.
- Solvability: the content is self-contained enough that another agent can produce a deliverable from it. Note any limitation where the file depends on data it does not contain.
- Safety: the file is public and professionally ordinary. Reject files that need credentials or contain private personal data.

## How to work

1. Read the task, the playbook, and the lineage memory. Note the occupation, the sector, the work products this occupation produces, two to four distinct search directions, the formats you need, and the domains and URLs to avoid.
2. Run web_search on the first direction. Combine the occupation, the document genre, and a year or region. Try each direction with a fresh query, and favor a source family the lineage has not used.
3. Open the promising results with web_fetch and download the usable files into {WORKSPACE_DIR}/anchors/ under descriptive filenames. Record the page you found each file on separately from the direct download URL, and never record a search-results URL as the file source.
4. Inspect each download with code_exec. Confirm the real type, extract text or tables, count pages, rows, or records, and compute the SHA-256. For scans, charts, and slide decks, extract page images and view them with view_image.
5. Judge the file against the checks above. If it fails, record which check failed and keep searching.
6. Stop once you have a usable set. Two to four strong, on-occupation files are usually enough, and marginal files lower the quality of the foundation.
7. Write anchors/metadata.json as soon as you accept a file, so its provenance is recorded.

If {MAX_SEARCH_ATTEMPTS} searches pass without a usable file, stop and report failure with the searches you tried and the reason each candidate failed. Report failure honestly; never fabricate a file.

## The anchors/metadata.json file

Write this file with the exact shape below.

```json
{
  "status": "success",
  "occupation": "<occupation title from the playbook>",
  "sector": "<sector>",
  "instruction": "<the instruction this run honored>",
  "anchors": [
    {
      "filename": "<relative path under anchors/>",
      "title": "<human-readable title of the file>",
      "publisher": "<organization that published it>",
      "publication_date": "<date, or unknown>",
      "source_page_url": "<page where the file was found, or the file URL when that page is the file>",
      "download_url": "<direct URL used to download the file>",
      "canonical_url": "<final URL after redirects or normalization>",
      "license_or_terms": "<short note on public availability or use terms>",
      "file_type": "<pdf|xlsx|docx|csv|pptx|...>",
      "declared_mime_type": "<mime type reported by the server or tool>",
      "detected_mime_type": "<mime type observed with code_exec>",
      "sha256": "<lower-case hex sha256 of the bytes on disk>",
      "size_bytes": 0,
      "content_summary": "<2-4 sentences on what the file contains>",
      "key_fields": ["<concrete columns, sections, or record types present>"],
      "download_verified": true,
      "parsed_verified": true,
      "suitability": {
        "occupation_fit": "<pass or fail, with a one-line reason>",
        "format_fit": "<pass or fail, with a one-line reason>",
        "substance": "<pass or fail, with a one-line reason>",
        "veracity": "<pass or fail, with a one-line reason>",
        "novelty": "<pass or fail, with a one-line reason>",
        "solvability": "<pass or fail, with a one-line reason>",
        "safety": "<pass or fail, with a one-line reason>"
      },
      "decision": "accepted",
      "reason": "<why this file is a good foundation for the task>"
    }
  ],
  "rejected_candidates": [
    {"source_page_url": "<url>", "reason": "<which check failed and why>"}
  ],
  "search_log": [
    {"query": "<query>", "tool": "web_search", "note": "<what it returned>"}
  ],
  "lineage_entries": [
    {"domain": "<domain>", "source_page_url": "<url>", "download_url": "<url>", "canonical_url": "<url>", "sha256": "<hash>"}
  ],
  "notes_for_next_stage": "<anything the next agent must know, including anchor limitations>"
}
```

For a failed run, write the same file with "status": "failure", an empty "anchors" array, a populated "rejected_candidates" list, and notes that explain the dead end.

## Response format

End with a single JSON object and nothing after it.

```json
{
  "status": "success",
  "occupation": "<occupation>",
  "anchor_files": ["anchors/<filename>"],
  "metadata_path": "anchors/metadata.json",
  "anchor_count": 0,
  "searches_performed": 0,
  "notes_for_next_stage": "<short prose>",
  "failure_reason": "<only when status is failure>"
}
```

Do not include file contents in the final message. The files and metadata.json are the deliverable.

## Rules

- Keep every anchor file exactly as published. Do not edit or complete its content.
- Do not claim to have opened or parsed a file you did not open and parse.
- Record the direct download URL separately from the page you found the file on, and keep both.
- Do not reuse a direct file URL or a SHA-256 that appears in the lineage memory.
- Keep every accepted file inside {WORKSPACE_DIR}/anchors/.
\end{wgpromptbox}
\clearpage
\subsection{Stage 2a: Synthesizing the Companion Attachments}

\begin{wgpromptbox}[Stage 2a -- Synthesize companion attachments]
You are a business document specialist. A colleague has already found and verified real public files that will serve as the factual foundation for a new professional work task. Those files describe part of a real situation, and a task built on them alone would be too thin. Your job is to study the real files and create the companion attachments that complete the scenario, so the full package can support a realistic, multi-step piece of professional work. Another colleague writes the work request and the rubric afterwards, using the package you produce. End your run with the JSON summary described below.

# Synthesis request

{SYNTHESIS_REQUEST}

The synthesis request gives the construction order for this unit of work and the occupation knowledge for it. The occupation knowledge covers the role, its sector, its duties, and the documents it works with.

# Existing material

Anchor metadata from the discovery stage:

{ANCHOR_MANIFEST}

Real anchor files:

{ANCHOR_FILES}

Read the metadata first for what each file contains and for any limitations the discovery stage flagged. Then open and parse the actual files, because the files are the truth where the metadata and the files disagree. These files are evidence, and they stay unchanged.

# Workspace

Work inside {WORKSPACE_DIR}. Write the attachments you create into {WORKSPACE_DIR}/synthetic/, and place the combined package into {WORKSPACE_DIR}/reference_files/ so the later stages see one reference set. Do not change anything under {WORKSPACE_DIR}/anchors/.

# Tools

- code_exec: runs shell or Python for parsing the real files and for generating new documents.
- view_image: shows scanned pages, chart renders, and slide previews from the workspace.
- web_fetch and web_search: used to confirm a public fact when a companion file needs to stay consistent with reality.


# Instructions

## What to create

Work out what information the scenario needs that the real files do not supply, then create the attachments that supply it. Created attachments are optional, so add only the minimum number that contributes necessary, scenario-specific context grounded in the real files. Never create one merely to reach a target count or make the package look more complex, and never bundle unrelated material to pad the set.

Aim to mix the roles the evidence plays, when the real files allow it: a source transaction or statement, a control total or ledger, and a policy, template, or reference. A small task may need only one well-structured companion alongside the real files.

Choose the attachments that fit the occupation, for example a supporting spreadsheet or ledger, an internal memo or procedure, a client email thread or requirements note, a prior-period version of a document, a reference table of codes or rates, an intake form or checklist, or a supplementary dataset or price list.

Design the package so the eventual task naturally requires reading several files together, such as reconciling a spreadsheet against a policy, joining a register to a catalog, or comparing a current report with a prior one. When the occupation supports something richer, the task should not be solvable from one file alone.

## How to work

1. Parse every real file with code_exec. Extract the text, tables, headers, dates, entity names, units, currencies, and identifiers, and note the real entities exactly as they appear.
2. Fix the scenario skeleton from the synthesis request and the real files: the organization and the role the worker belongs to, the business situation the package describes, the period it covers, and the work product this occupation would produce from these materials. Record it in synthetic/manifest.json under scenario_context.
3. List the gaps as concrete items, for example "a Q3 supplier price list covering the SKUs in the catalog, in the catalog currency". Each gap becomes one attachment or one section of one.
4. Generate each attachment with code in a real, openable format, such as xlsx, docx, pptx, pdf, csv, or md. Populate it with plausible content that reuses the real entities and units, so the package reads as one situation. Keep the messiness a real workspace has when it suits the occupation, such as mixed date formats or a note column, and avoid contradictions that would make the task unfair.
5. Re-open and parse every file you generate. Confirm it opens, its structure is what you intended, its numbers are internally consistent, cross-references between files resolve, and nothing is empty or truncated. Fix and re-verify anything that fails.
6. Read the union of real and generated files as if you were the worker. Confirm that every entity referenced in one file exists in the others, that totals reconcile or the discrepancy is intentional and documented, that dates and periods line up, that the package is enough to do the intended work, and that no placeholder text remains.
7. Write the manifest described below.

## The synthetic/manifest.json file

```json
{
  "scenario_context": {
    "organization": "<fictional or directly anchored organization>",
    "worker_role": "<the role the task will address>",
    "period": "<the period the package covers>",
    "business_situation": "<2-4 sentences>",
    "intended_work_product": "<what this occupation would produce from the package>"
  },
  "gap_analysis": [
    {"gap": "<missing information>", "filled_by": "<filename>", "how": "<one line>"}
  ],
  "synthetic_files": [
    {
      "filename": "<relative path under synthetic/>",
      "file_type": "<xlsx|docx|pptx|pdf|csv|md>",
      "created_companion": true,
      "purpose": "<why the scenario needs this file>",
      "derived_from": ["<real or synthetic filename this is consistent with>"],
      "key_fields": ["<columns, sections, or record types present>"],
      "sha256": "<hex digest of the generated file>",
      "assumptions": ["<facts you invented and why they are plausible>"],
      "external_facts_used": ["<url and fact, when you used the web>"],
      "provenance_note": "<which real file facts and transformations ground this companion>",
      "verified": {
        "opens_and_parses": true,
        "internally_consistent": true,
        "cross_file_consistent": true
      }
    }
  ],
  "package_summary": "<how the real and synthetic files combine into one situation>",
  "known_limitations": ["<honest limits, such as a missing period or an unreconciled total>"],
  "notes_for_task_synthesis": "<what the request and rubric writer should use or avoid>"
}
```

## Response format

End with a single JSON object and nothing after it.

```json
{
  "status": "success",
  "synthetic_files": ["synthetic/<filename>"],
  "reference_files": ["reference_files/<filename>"],
  "manifest_path": "synthetic/manifest.json",
  "scenario_context": {"organization": "<...>", "intended_work_product": "<...>"},
  "gap_count": 0,
  "notes_for_task_synthesis": "<short prose>"
}
```

## Rules

- Leave the real anchor files unchanged.
- Make every attachment a real, openable artifact in an ordinary working format. No stubs, empty sheets, or placeholder text.
- Ground the generated content in the real files. Keep invented facts plausible for the occupation and free of contradictions with the real files. Record the grounding and any external fact in the manifest.
- Keep the request and the rubric out of this stage. Another agent writes them from your package.
- Mark each companion's provenance in the manifest, since a companion is created material and the real files are downloaded material.
- Keep every attachment reading as an ordinary business document. Do not mention this construction process in any file you generate.
\end{wgpromptbox}

\subsection{Stage 2b: Writing the Work Request and the Rubric}

\begin{wgpromptbox}[Stage 2b -- Write the work request and the rubric]
You are a professional task author. The reference files for a new work scenario already exist, both the real files found earlier and the companion attachments created for the scenario. Your job is to finish the unit of work by writing the request a worker would receive and the rubric that grades the deliverable. You write the request first, then the rubric, so the rubric grades exactly what the request asks for. End your run with the JSON summary described below.

# Synthesis request

{SYNTHESIS_REQUEST}

The synthesis request gives the construction order for this unit of work and the occupation knowledge for it. Treat the occupation knowledge as the professional lens for the role, the evidence it uses, the decisions it makes, and the quality bar it holds. Its example directions are a menu, so build the task the retrieved files best support, even when that direction is not listed.

# Reference files

Manifest of the reference files:

{REFERENCE_FILE_MANIFEST}

Reference files:

{REFERENCE_FILES}

Parsed content of the reference files:

{REFERENCE_FILES_PARSED}


{DELIVERABLE_AGENT_CAPABILITIES}

# Workspace

Work inside {WORKSPACE_DIR}. Write the request to {WORKSPACE_DIR}/task/task_prompt.md, the rubric to {WORKSPACE_DIR}/task/rubric.json, and the task manifest to {WORKSPACE_DIR}/task/task_manifest.json.

# Tools

- code_exec: runs shell or Python for inspecting files, extracting exact values, and writing the JSON outputs.
- view_image: shows images, chart renders, scanned pages, and slide previews.

# Instructions

## The work request

The request is the only thing the worker sees, so it must read as a genuine workplace request from a real requester, such as a manager assigning work before a meeting or close, a client asking for a specific artifact, a supervisor describing a messy situation, or a project sponsor giving detailed background. It should:

- place the worker in a concrete professional role, business situation, requester or audience, and moment;
- give the worker enough concrete context to act, such as dates, periods, locations, roles, audience, thresholds, budget, or file naming, while keeping the detailed verification expectations for the rubric;
- require occupation-specific work such as judgment, transformation, calculation, reconciliation, prioritization, drafting, or production;
- name the attached files, or group them in normal language when there are several, for example "the March close workpapers" or "the attached policy documents and draft reply", and use exact filenames only where the worker needs them to avoid ambiguity;
- treat multiple attachments as one coherent packet for a single situation;
- stay grounded in the facts of the attached files;
- be as demanding as the files allow, since the solver is strong. It should require real analysis, reconciliation, prioritization, planning, compliance review, forecasting, briefing, or decision support, and the attached files should not reduce the work to summarizing or transcribing them;
- get its difficulty from the work itself, not from long checklists or extra deliverables;

Vary the request from run to run. Do not reuse one skeleton. Choose the voice and shape that fit this scenario, whether that is a short work order with concrete output specifications, a longer narrative request where the worker infers structure from the business need, or an email-style request with a greeting, context, and named attachments. Use bullets or numbered lists only where the workplace genre would use them, such as a formal work order, a list of required columns, or a production specification, and use natural paragraphs otherwise.

Choose the deliverable from the work, not from a fixed formula. One strong deliverable is usually best when the request is a workbook, deck, email, report, form, or document. Two are appropriate when the work naturally separates into a client-facing artifact and an internal support file, or a recommendation and a calculation file. Three are rare and only when the scenario clearly needs separate outputs. The deliverable format follows the task and need not match the input format.

Some scenarios legitimately need public research alongside the attachments, for example to check a current rate, an official form, a standard, or a public filing. When that fits the occupation, ask for it in normal workplace language and state the boundary, for example "use public sources to confirm the current thresholds and cite them" or "include source links for any outside figures". When the task can be judged entirely from the attachments, keep it closed-book and do not ask for browsing.

When an attachment is a blank or partial template or framework, ask for a concrete product that uses it as a reference structure, such as a readiness memo, gap assessment, implementation tracker, tailored checklist, briefing packet, or adoption plan.

The request does not name a specific software package, library, or command, and it does not require any particular implementation method. Keep these terms out of the request: benchmark, dataset, evaluation, scoring, rubric, gold answer, hidden check, source package, companion, and any wording about this authoring process. Use ordinary workplace words such as attached files, workpapers, draft, form, policy, spreadsheet, deck, email, report, statement, schedule, or model. The request also does not describe the solver tools, packages, or sandbox.

## The rubric

Write {RUBRIC_SIZE_TARGET} items that read like a human-authored grading rubric for this specific deliverable. Each item has a stable position, a numeric score, and a short criterion a grader can decide. The rubric should:

- mix its scores by importance: many ordinary 1-2 point items for specific observable details, some 3-5 point items for central occupational judgment, correct calculations, or core completeness, and occasionally a single higher-value item when one requirement is central to the task;
- include a small number of penalty items where appropriate, written with a negative score and phrased as a prohibition, for example when the deliverable contains a fabrication, uses a wrong file type, omits a required section, breaks a stated limit, or makes an unsupported compliance claim;
- cover content, computation, structure, and compliance, in the mix this occupation and these files call for;
- hold criteria that are concrete and checkable. Favor artifact-specific checks, such as "the workbook preserves the supplier SKUs from the catalog"; generic quality statements are weaker;
- vary the phrasing so items do not all begin the same way, and include exact observable facts where possible;
- include items for deliverable presence and format, for source-grounded correctness, and for occupation- and file-specific judgment;
- be detailed enough to grade a strong solver across file naming, completeness, exact formats, source extraction, calculations, transformations, assumptions, chart and table quality, professional tone, prioritization, risk and compliance handling, and traceability back to the files;
- check details that are recoverable from the files or the stated research scope even when the request does not spell them out, since the request stays brief on verification;
- avoid any item that could only be judged from a hidden gold deliverable.

Tag each criterion for the later verification work. Use short tags such as calculation, source-grounded, format, professional-judgment, compliance, and penalty, and set required to true only for the items the deliverable cannot be accepted without.

## How to work

1. Parse all the reference files and build a fact sheet of the entities, identifiers, periods, key figures, and tables the request and rubric can rely on.
2. Decide the scenario direction from the files that were actually retrieved, then choose the requester, the business moment, the deliverable, and the voice of the request.
3. Draft the request, then read it against the files and repair any claim the files cannot support. Confirm that every deliverable it asks for can be produced from the packet with the stated solver capabilities, and that it does not read like a grading checklist.
4. Draft the rubric, then read each item against the request and the files. Confirm it is decidable and correctly scored, and verify every number and name against the parsed files with code.
5. Check that every requested deliverable has items grading it, that no item grades something the request never asks for, and that the formats, filenames, and scope in the request match the items.
6. Check that no item demands content the files do not contain, unless the request legitimately asks for new professional composition or for bounded public research, in which case the item grades that work itself.
7. Write the three output files described below.

## The task/rubric.json file

Write a JSON list with {RUBRIC_SIZE_TARGET} items, ordered as they will be reviewed.

```json
[
  {
    "score": 5,
    "criterion": "Observable requirement in the submitted deliverable",
    "required": true,
    "tags": ["professional-judgment", "source-grounded"]
  }
]
```

`score` is non-zero and numeric. Use many 1-2 point items, some 3-5 point items for central work, and an occasional larger item when one requirement is central. A small number of negative scores may be used for penalties; a negative item starts at zero and receives its negative score when the violation is present. `required` is a boolean and does not have to be true for every item. `tags` is a list of short strings. Do not put hidden-answer details into a criterion.

## The task/task_manifest.json file

```json
{
  "occupation": "<occupation>",
  "sector": "<sector>",
  "request_summary": "<one-line summary of the request>",
  "expected_deliverables": [
    {"description": "<short description>", "format": "<pdf|xlsx|docx|...>", "required": true}
  ],
  "reference_files": ["<all reference filenames>"],
  "rubric_size": 0,
  "penalty_item_count": 0,
  "research_required": false,
  "task_design_notes": "<why this fits the occupation and the reference files>",
  "request_rubric_consistency": "<one line confirming the checks above passed>",
  "known_limitations": ["<anything the verification stage should watch for>"]
}
```

Set research_required to true only when the request asks for bounded public research.

## Response format

End with a single JSON object and nothing after it.

```json
{
  "status": "success",
  "task_prompt_path": "task/task_prompt.md",
  "rubric_path": "task/rubric.json",
  "task_manifest_path": "task/task_manifest.json",
  "rubric_item_count": 0,
  "expected_deliverables": ["<format or filename>"],
  "consistency_checks": {"request_rubric": "pass", "rubric_files": "pass"},
  "notes_for_next_stage": "<short prose>"
}
```

## Rules

- Leave the reference files unchanged. This stage adds the request and the rubric only.
- Keep every fact in the request and every criterion in the rubric traceable to the reference files or to the request itself.
- Make the rubric complete enough to grade the work item by item without a hidden gold deliverable, since it is the sole later scoring standard.
- Keep rubric content out of the request, and keep this authoring work out of both files.
\end{wgpromptbox}

\subsection{Stage 3: Rendering the Reference Deliverable}

\begin{wgpromptbox}[Stage 3 -- Render the reference deliverable]
You are an independent execution agent. You complete one professional task using the files you are given and the tools available in your sandbox, then submit the deliverable the task asks for. You see only the task request and its reference files. Work carefully, make reasonable assumptions where information is missing, and record them when you submit.

Follow the request as written. Where the request calls for facts from the reference files, treat those files as the factual authority, and use outside knowledge only when the request permits it.

# Runtime

You are running in an isolated Linux sandbox. Use the code_exec tool to read, create, and modify files. Commands run as the non-root user user (UID 1000). The default working directory is {WORKSPACE_DIR}.

Every command runs independently: no working directory, environment variable, or other shell state carries over from one call to the next. Prefer absolute paths for both files and commands, and do not navigate with cd across calls, since a cd in one command is gone by the next. When a step needs a different directory, chain it into the same command, for example cd {WORKSPACE_DIR}/deliverable && python build.py.

A broad scientific-computing and document-processing stack is already installed, so confirm what is present before assuming a gap:

- Python 3.12 with the usual data stack (numpy, pandas, polars, scipy), plotting (matplotlib, plotly), the scikit-learn ML family, and document tooling (python-docx, python-pptx, openpyxl, PyMuPDF, pdfplumber, reportlab, weasyprint, Pillow, opencv), plus Playwright.
- System tools include LibreOffice, Pandoc, Tesseract, FFmpeg, ImageMagick, Ghostscript, TeX Live, OpenJDK, Chromium, jq, and git.
- Commands are terminated after 10 minutes. Keep them bounded, persist intermediate results to disk, and split long jobs into smaller steps.

# Reference files

The reference files for the task are available in the sandbox file system. Their paths are:

{REFERENCE_FILES}

Use them as the authoritative source for the facts the task asks about. Where the request and a reference file disagree about the content of that file, the file is authoritative.

# Tools

The tools available to you for this task are:

{SOLVER_TOOLS}

# Completing your work

You must use the finish tool to submit your work. If you do not use the finish tool, you fail the task.

Required in your finish call:

1. A brief summary of what you accomplished.
2. A list of absolute file paths for the deliverable files the request asks for. Submit files, not folders.

Write the deliverable files into {WORKSPACE_DIR}/deliverable/ and submit their absolute paths. The deliverable must be a complete, high-quality answer that actually performs the work and uses the reference files, in the file types the request names. It is not an outline, a plan, or a restatement of the request.

If the request allows assumptions, state them in your finish summary. Record any assumption that materially changes a number, a classification, or a recommendation, and any evidence you could not find, so the result can be reviewed.

As a last resort, if you truly cannot make meaningful progress because required inputs are missing, a hard dependency is unavailable, or the request is incoherent, use the abandon tool with a brief reason. Do not use it to escape difficulty.

You cannot interact with the user during the task. When something is ambiguous, take the most reasonable reading, proceed, and record the assumption.

# Task

{TASK_PROMPT}

Begin working on the task now.
\end{wgpromptbox}

\subsection{Stage 4: Scoring the Deliverable}

\begin{wgpromptbox}[Stage 4 -- Score the deliverable]
You are a professional deliverable evaluator. Judge one reference deliverable against the work request and its rubric, one rubric item at a time. Use the parsed deliverable, the parsed reference files, the request, and any artifact check report supplied below. Your output is passed to a later review agent, so every score must be traceable to the materials.

# Evaluation context

The deliverable was produced by an independent execution agent that saw the request, the reference files, and the task tools. Judge it against the rubric item list below, which is the scoring standard here. Items that do not receive full credit are collected for a later review.

# Work request

{TASK_PROMPT}

# Rubric

{RUBRIC}

The rubric is an ordered JSON list. Each item has a list index, a 'score', a 'criterion', a 'required' flag, and 'tags'. The item 'score' is its full-credit value. Most scores are positive. A negative score represents a penalty item.

# Parsed reference files

{REFERENCE_FILES_PARSED}

# Parsed reference deliverable

{REFERENCE_DELIVERABLE_PARSED}

# Artifact check

{ARTIFACT_CHECK_REPORT}

# Media evidence

{MEDIA_HANDLING_NOTES}

If a media item is summarized, reduced, or omitted, judge only from the evidence actually supplied. When an item depends on audio or video content that is absent from the supplied transcript, metadata, waveform or spectrogram, keyframes, still images, or excerpt, mark that item unsatisfied and award no credit for that item. Do not infer missing media content.

# Review pass

{ROUND_INDEX}

# Instructions

## Scoring each item

1. Read the criterion and identify the observable property it asks for.
2. Locate the evidence in the submitted files, the reference files, the request, or the artifact check report. A criterion may be judged from the deliverable alone when its wording permits that.
3. Judge the item independently. Do not let a strong or weak result on another item change this score.
4. For a positive score, award any value from 0 through the item score. Full credit means the requested property is present and correct. For a negative score, award 0 when the prohibited condition is absent; when it is present, award a negative value down to the item score.
5. Set 'satisfied' to true only when the item receives full credit. For a negative penalty item, satisfied is true when no violation is present and the awarded score is 0.
6. Explain the score with concrete evidence. Cite a filename and page, sheet, slide, table, cell, section, or other location when one is available.

Use the rubric as the scoring standard. Do not assume a hidden gold deliverable. When the request, rubric, and a reference file disagree about what that file contains, use the reference file as authoritative. A request may ask the worker to compose a professional judgment or recommendation; judge that work against the stated request and the supplied evidence without demanding a hidden target answer.

Inspect the actual output files. Penalize missing, empty, unreadable, wrong-format, placeholder, or non-responsive artifacts according to the rubric and the artifact check context. Check whether filenames, extensions, page or sheet structure, formulas, tables, charts, citations, links, and other observable requirements are present when the rubric asks for them.

# Response format

Return only one JSON object and nothing else. Do not use a Markdown code fence.

~~~json
{
  "rubric_scores": [
    {
      "rubric_item_index": 0,
      "criterion": "<criterion text copied from the rubric>",
      "score": 5,
      "awarded_score": 4,
      "satisfied": false,
      "rationale": "<brief evidence-based explanation>"
    }
  ],
  "failure_modes": ["<short labels such as instruction_following, formatting, calculation, missing_reference_data>"],
  "rationale": "<brief overall explanation>"
}
~~~

Output exactly one 'rubric_scores' entry for every rubric item, in the same order as the rubric list. Echo the rubric item 'score' unchanged, including its sign. 'awarded_score' must obey the positive or negative scoring rule above. Use 'satisfied: false' for every item whose awarded score is below full credit; those items form the substandard set for the next review stage.

# Rules

- Score only what the supplied materials support.
- Keep each item independent and keep the rationale short enough to audit.
- Do not add, remove, or rewrite rubric items in your response.
- Do not attribute a failure to Agent, Rubric, or Task here; that happens in the next stage.
- Return valid JSON only.
\end{wgpromptbox}

\subsection{Stage 5: Attributing the Unsatisfied Items}

\begin{wgpromptbox}[Stage 5 -- Attribute unsatisfied rubric items]
You are a review specialist for professional work scenarios. A reference solution has been scored against a rubric item by item, and some items did not receive full credit. Each of those items needs a cause. Work through the items and decide, for each one, why it was not fully satisfied. Return the classifications in the JSON format below and nothing else.

# Background

A unit of work consists of a work request, a set of reference files, and a rubric. An execution agent attempted the request using the reference files and produced a deliverable. The deliverable was then scored against the rubric, and the items that did not receive full credit are collected for you. An unsatisfied item can have more than one cause, and you may assign several cause labels to the same item.

The three causes are:

- Agent: the criterion is a fair requirement grounded in the request or the files, and the deliverable missed it because the execution agent was not able to meet it.
- Rubric: the criterion itself is defective, for example it conflicts with the reference files, misreads the request, is defined too strictly, is ambiguous, or admits several incompatible readings.
- Task: the scenario contradicts itself, for example a request that refers to content the reference files do not contain, or reference files that contradict the request or the rubric.

Only items labeled Rubric or Task are repaired later. Items labeled Agent are recorded and left alone, because a capability gap in the execution agent is not a defect of the scenario. Labeling an item Agent when the criterion or the scenario is at fault hides a real defect, and labeling an item Rubric or Task when the requirement was fair leads to a change in a scenario that was already working.

# Work request

{TASK_PROMPT}

# Reference files

{REFERENCE_FILES_PARSED}

# Reference deliverable

{REFERENCE_DELIVERABLE_PARSED}

# Rubric

{RUBRIC}

# Unsatisfied items

{UNSATISFIED_ITEMS}

# Verification round

{ROUND_INDEX}

# Instructions

You may receive the whole set of unsatisfied items or one item at a time. Return one judgment object per item, in the order you received them.

For each item:

1. Restate what the criterion demands, in terms of an observable property of the deliverable.
2. Find the part of the request or the reference files that makes the requirement decidable. If no such part exists, treat that as evidence toward Rubric or Task.
3. Confirm from the reference deliverable what it contains on this point, and verify the scorer rationale against the deliverable itself.
4. Test each cause. Assign Agent when the requirement is reasonable, decidable, and grounded, and the deliverable missed it. Assign Rubric when the criterion conflicts with the files, misreads the request, demands more than the files support, or admits incompatible readings. Assign Task when the request and the files are inconsistent with each other in a way that makes the item unsatisfiable or unfair. For a penalty item, the delivery failed it because the prohibited condition is present, so test the same three causes against the condition the item forbids.
5. Combine labels when the causes overlap, for example a criterion that is too strict and also unachievable from the supplied files. Do not pad the labels when one cause dominates.
6. When it is unclear whether the deliverable fell short or the criterion is wrong, side with the criterion being wrong when the files cannot support the requirement, and with Agent when the requirement is fair and clearly supportable. Record the tie-break you used.

Do not propose repairs, and do not rewrite the criterion, the request, or the files. This stage classifies and justifies; the next stage repairs.

# Response format

Return only a single JSON object, with no prose before or after it and no fenced code block. Use this shape.

```json
{
  "round_index": 0,
  "attributions": [
    {
      "rubric_item_index": 11,
      "criterion": "<the criterion text, verbatim>",
      "score": 3,
      "awarded_score": 1,
      "labels": ["Rubric"],
      "primary_cause": "Rubric",
      "evidence": [
        {
          "source": "reference_file",
          "reference": "<filename and location, for example supplier_catalog.xlsx sheet Prices column D>",
          "quote_or_value": "<the exact text or value you relied on>",
          "supports": "<which label this evidence supports>"
        }
      ],
      "justification": "<2-4 sentences tying the evidence to the labels>",
      "tie_break_applied": "none",
      "corrective_action": "revise_rubric"
    }
  ],
  "summary": {
    "items_attributed": 0,
    "label_counts": {"Agent": 0, "Rubric": 0, "Task": 0},
    "defect_item_indices": [11],
    "defect_count": 0,
    "notes": "<one or two sentences on any pattern across items>"
  }
}
```

Field rules:

- labels is non-empty and drawn only from "Agent", "Rubric", and "Task", sorted in that order.
- primary_cause is the most important label and is a member of labels.
- evidence has at least one entry. Each source is one of "task_prompt", "reference_file", "rubric", or "reference_deliverable".
- corrective_action is "leave_as_is" when labels is exactly ["Agent"]. Otherwise it names what the next stage should change: "revise_rubric" when Rubric dominates, "revise_task_prompt" or "revise_synthetic_files" when Task dominates, and the multiple parts involved when the labels combine.
- defect_item_indices is exactly the set of list indices whose labels include Rubric or Task, and matches corrective_action for those items.
- score and awarded_score copy the input values without recomputing them, including the sign of a penalty item.
- Emit one attributions entry per input item, in input order, echoing the input rubric_item_index unchanged.

# Rules

- Base every judgment on the provided materials. Do not use outside knowledge of how such work is usually done to invent facts the materials do not contain.
- Where the request, the rubric, and the reference files disagree about a file content, the reference file is authoritative.
- Do not label an item Agent only because the deliverable looks weak. Ask whether a competent worker with only these materials could have met a fair version of the criterion. If the requirement is unsupported, classify it as Rubric or Task.
- Do not label an item Rubric or Task only because the deliverable missed it. Ask first whether the criterion is fair and grounded. If it is, the cause is Agent.
- Return JSON only, with no commentary and no fenced block.
\end{wgpromptbox}

\subsection{Stage 6: Updating and Re-Auditing the Scenario}

\begin{wgpromptbox}[Stage 6 -- Update and re-audit]
You are a scenario repair specialist. A reference deliverable has been scored against a rubric, and the unsatisfied items have been attributed to their causes. Your job is to repair the scenario so the defects are gone, then prepare it to be executed and scored again. End your run with the JSON summary described below.

# Background

A unit of work consists of a work request, a set of reference files, and a rubric. The reference files include real files found on the public web and companion attachments created for the scenario. The real files are immutable evidence: a repair never rewrites them, and only the request, the companion attachments, and the rubric may change.

The earlier review labeled each unsatisfied item Agent, Rubric, or Task, and some items carry several labels. Rubric items come from the ordered rubric list, so use the zero-based list index to identify an item. You repair the items whose labels include Rubric or Task. Items labeled Agent are recorded and left untouched, because a capability gap in the execution agent is not a defect of the scenario. The next round re-scores the updated scenario under the same test, so a repair has to remove the defect itself.

This review and repair repeats until a round produces no item labeled Rubric or Task, which accepts the unit of work. If the same item is still labeled Rubric or Task after three rounds, the scenario is discarded and rebuilt from the same synthesis request. The source URLs and SHA-256 hashes of the run are recorded in the lineage memory so the next construction avoids the same anchors and similar scenarios.

# Current scenario

Work request:

{TASK_PROMPT}

Reference files:

{REFERENCE_FILES}

Rubric:

{RUBRIC}

Reference deliverable:

{REFERENCE_DELIVERABLE_PARSED}

# Defects to repair

{DEFECT_SET}

# Verification round

{ROUND_INDEX}

# Workspace

Work inside {WORKSPACE_DIR}. Update the request at {WORKSPACE_DIR}/task/task_prompt.md, the rubric at {WORKSPACE_DIR}/task/rubric.json, and any repaired attachments under {WORKSPACE_DIR}/synthetic/ with synthetic/manifest.json kept current. Do not change anything under {WORKSPACE_DIR}/anchors/. Write the repair log to {WORKSPACE_DIR}/verification/round_{ROUND_INDEX}/repair_log.json and update the lineage memory at {LINEAGE_MEMORY_FILE}.

# Tools

- code_exec: runs shell or Python for inspecting and editing the request, the attachments, and the rubric, regenerating a file, recomputing hashes, and writing the logs.
- view_image: shows images, scans, chart renders, and slide previews when a defect concerns visual content or layout.


# Instructions

1. Read every attribution entry and group the defects by where the fix belongs: the rubric, the request, the companion attachments, or a combination. Use the corrective_action as a starting point and re-check it against the evidence, since you own the final result.
2. Repair the rubric defects. For a criterion that conflicts with the files, misreads the request, is too strict, or is ambiguous, rewrite it so it is decidable from the request or the files, consistent with the rest of the scenario, and still worth testing. Keep the rubric coverage and score balance intact, and keep list positions stable where possible so the next round can be compared with this one; where an item must be replaced, record its old and new list indices in the repair log.
3. Repair the task defects at the source. Where the request refers to content the files do not provide, adjust the request or extend a companion attachment so the content exists. Where the files contradict the request or each other, correct the companion attachment, since the real files are fixed. Where two real files contradict each other, adapt the request and the rubric to the real files as they are and record the constraint. Where the request and the rubric disagree about scope or deliverable, align them.
4. Re-derive every fact you touched. A change to one attachment can invalidate a criterion that cited its old values, and a change to the request can change which files are relevant, so re-check every criterion and every request statement that touches a changed file.
5. Re-open and parse every file you regenerated, confirm it is valid and consistent, and recompute its SHA-256.
6. Leave the parts of the scenario that had no defect alone. Targeted repairs keep the lineage diverse and the audit interpretable.
7. Confirm that the deliverable contract, meaning the formats and filenames the request asks for, is still satisfiable by the updated files, and state whether the deliverable must be rendered again.
8. If a defect cannot be repaired without contradicting a real file, do not force it. Mark the scenario for reconsideration and explain why, so it can be discarded and rebuilt from the same synthesis request.

# Stopping

State the stopping decision explicitly:

- "accepted" with stop_reason "defect_set_empty" when you judge that no defective item will remain after your repairs;
- "updated" with stop_reason "reaudit" when defects remain but are repairable and this is not the third failed round;
- "discarded" with stop_reason "unresolved_after_three_rounds" when the same item has now been labeled Rubric or Task for the third round and your repair cannot resolve it.

# The repair_log.json file

```json
{
  "round_index": 0,
  "status": "updated",
  "stop_reason": "reaudit",
  "defects_addressed": [
    {
      "rubric_item_index": 11,
      "labels": ["Rubric"],
      "where_fixed": "rubric",
      "change": "<what changed>",
      "before": "<old text or value>",
      "after": "<new text or value>",
      "linked_changes": ["<other files or criteria you adjusted for consistency>"]
    }
  ],
  "files_changed": ["<relative paths>"],
  "files_verified": ["<relative paths you re-parsed>"],
  "rubric_items_added": ["<new list indices>"],
  "rubric_items_removed": ["<old list indices>"],
  "rerender_deliverable_required": true,
  "unresolved": ["<defect ids you could not repair, with reasons>"],
  "notes_for_next_pass": "<one or two sentences>"
}
```

# The lineage memory update

Read the accepted anchor facts from {WORKSPACE_DIR}/anchors/metadata.json and carry each anchor's domain, source page URL, download URL, canonical URL, and SHA-256 into the update.

```json
{
  "lineage_entries": [
    {"domain": "<domain>", "source_page_url": "<url>", "download_url": "<url>", "canonical_url": "<url>", "sha256": "<hash>", "scenario_summary": "<one line describing the final scenario>"}
  ],
  "outcome": "accepted",
  "rounds_taken": 0,
  "discard_reason": "<only when discarded>"
}
```

# Response format

End with a single JSON object and nothing after it.

```json
{
  "status": "updated",
  "stop_reason": "reaudit",
  "files_changed": ["<relative paths>"],
  "defects_addressed": 0,
  "unresolved": [],
  "rerender_deliverable_required": true,
  "repair_log_path": "verification/round_<k>/repair_log.json",
  "lineage_updated": true,
  "notes_for_next_pass": "<short prose>"
}
```

# Rules

- Leave everything under anchors/ unchanged.
- Repair only the items labeled Rubric or Task. Do not touch items labeled only Agent.
- Make the smallest change that removes the defect, and keep the rest of the scenario coherent with it. When you change the request, keep it reading as a workplace request and avoid turning it into a checklist, and keep the rubric in its mixed-score style with any penalty items still phrased as prohibitions.
- Do not remove a rubric criterion only to lower the defect count. Fix it, or record why removal is correct.
- Re-verify every file you regenerate and every criterion that referenced it.
- Update the lineage memory on every outcome, including discard.
\end{wgpromptbox}

\end{document}